\documentclass[letterpaper, 10 pt, conference]{ieeeconf}  

\IEEEoverridecommandlockouts                              

\title{\LARGE \bf
	A Biologically Inspired Design Principle\\ for Building Robust Robotic Systems}

\usepackage{graphics} 
\usepackage{epsfig} 
\usepackage{mathptmx} 
\usepackage{times} 
\usepackage[cmex10]{amsmath}
\usepackage{amssymb}  
\usepackage{ifpdf}
\usepackage{xcolor}
\definecolor{sd}{gray}{0.5}
\usepackage{colortbl}
\usepackage{subcaption}
\usepackage{graphicx}
\usepackage{balance}
\usepackage[noend]{algorithmic}
\usepackage{framed}
\usepackage{tabularx}
\usepackage{gensymb}
\usepackage{todonotes}
\usepackage{array,multirow,graphicx}
\usepackage{diagbox}
\usepackage{booktabs}
\usepackage{siunitx}
\usepackage{tikz}
\usetikzlibrary{arrows.meta}
\usepackage{contour}
\usepackage{makecell}
\usepackage{multirow}
\usepackage[normalem]{ulem}
\usepackage{todonotes}
\usepackage{xspace}
\usepackage{array}
\usepackage{flushend}
\usepackage{cite}
\usepackage{float}
\usepackage{mathtools}
\usepackage{multirow}
\usepackage{booktabs}
\usepackage{hyperref}
\usepackage{cleveref}
\let\labelindent\relax
\usepackage{enumitem}
\hypersetup{
	colorlinks=true,
	urlcolor=blue
}
\usepackage{balance}
\usepackage{fancyhdr}
\usepackage{array}
\usepackage{flushend}  
\usepackage[ruled,vlined]{algorithm2e}

\newcommand\tparbox[2]{\protect\parbox[t]{#1}{\protect\raggedright #2}}

\usepackage{xcolor}
\definecolor{amber}{rgb}{1.0, 0.75, 0.0}

\newcommand{\extensionone}{\href{https://drive.google.com/file/d/1t7CinBnHtO9AQ0xCjCWga1voE8DOqBCU/view?usp=drive_link}{Extension 1}}
\newcommand{\extensiontwo}{\href{https://drive.google.com/file/d/1N-RuRst7JEkw0FkuU9fagq2-p_D2C2Ps/view?usp=drive_link}{Extension 2}}
\newcommand{\extensionthree}{\href{https://drive.google.com/file/d/1uL3gHLiOIdi5IMOsvAekdeRUrAKkBfXb/view?usp=drive_link}{Extension 3}}
\newcommand{\extensionfour}{\href{https://drive.google.com/file/d/1MBQbWD6URcwmpxKzxptegMXWPgKYo4yY/view?usp=drive_link}{Extension 4}}
\newcommand{\extensionfive}{\href{https://drive.google.com/file/d/1rtyCpIBigzdgLtRp8xuV-3S_4rdevsw5/view?usp=drive_link}{Extension 5}}

\newcommand{\baseSystem}{\textbf{Base}}

\newcommand{\weakPlanning}{\textbf{Base + \textcolor{red}{P}\textcolor{blue}{C}}}

\newcommand{\weakControlPlanning}{\textbf{Base + \textcolor{red}{P}\textcolor{blue}{C}\textcolor{amber}{P} + \textcolor{red}{P}\textcolor{blue}{C}}}

\newcommand{\weakPerceptionControl}{\textbf{Base + \textcolor{red}{P}\textcolor{blue}{C}\textcolor{amber}{P} + \textcolor{blue}{C}\textcolor{amber}{P}}}

\newcommand{\richInteractions}{\textbf{Base + \textcolor{red}{P}\textcolor{blue}{C}\textcolor{amber}{P} + \textcolor{red}{P}\textcolor{blue}{C} + \textcolor{blue}{C}\textcolor{amber}{P}}}

\author{Xing Li$^{^\star,1,2}$ \quad\quad Oussama Zenkri$^{^\star,1,2}$ \quad\quad Adrian Pfisterer$^{^\star,1,2}$ \quad\quad Oliver Brock$^{1,2}$ 
	\thanks{$^\star$ Equal contribution}
	\thanks{$^{1}$ Robotics and Biology Laboratory, Technische Universität Berlin}
	\thanks{$^{2}$ Science of Intelligence, Research Cluster of Excellence, Berlin}%
	\thanks{We gratefully acknowledge funded provided by the Deutsche Forschungsgemeinschaft (DFG, German Research Foundation) under Germany’s Excellence Strategy – EXC 2002/1 “Science of Intelligence” – project number 390523135.}
}

\begin{document}
	
	\maketitle
	\pagestyle{empty}

	\begin{abstract}
		
		
		Robustness, the ability of a system to maintain performance under significant and unanticipated environmental changes, is a critical property for robotic systems. While biological systems naturally exhibit robustness, there is no comprehensive understanding of how to achieve similar robustness in robotic systems. In this work, we draw inspirations from biological systems and propose a design principle that advocates active interconnections among system components to enhance robustness to environmental variations. We evaluate this design principle in a challenging long-horizon manipulation task: solving lockboxes. Our extensive simulated and real-world experiments demonstrate that we could enhance robustness against environmental changes by establishing active interconnections among system components without substantial changes in individual components. Our findings suggest that a systematic investigation of design principles in system building is necessary. It also advocates for interdisciplinary collaborations to explore and evaluate additional principles of biological robustness to advance the development of intelligent and adaptable robotic systems. 
		
	\end{abstract}
	
\section{Introduction}
\begin{figure*}
	\centering
	\includegraphics[width=\textwidth]{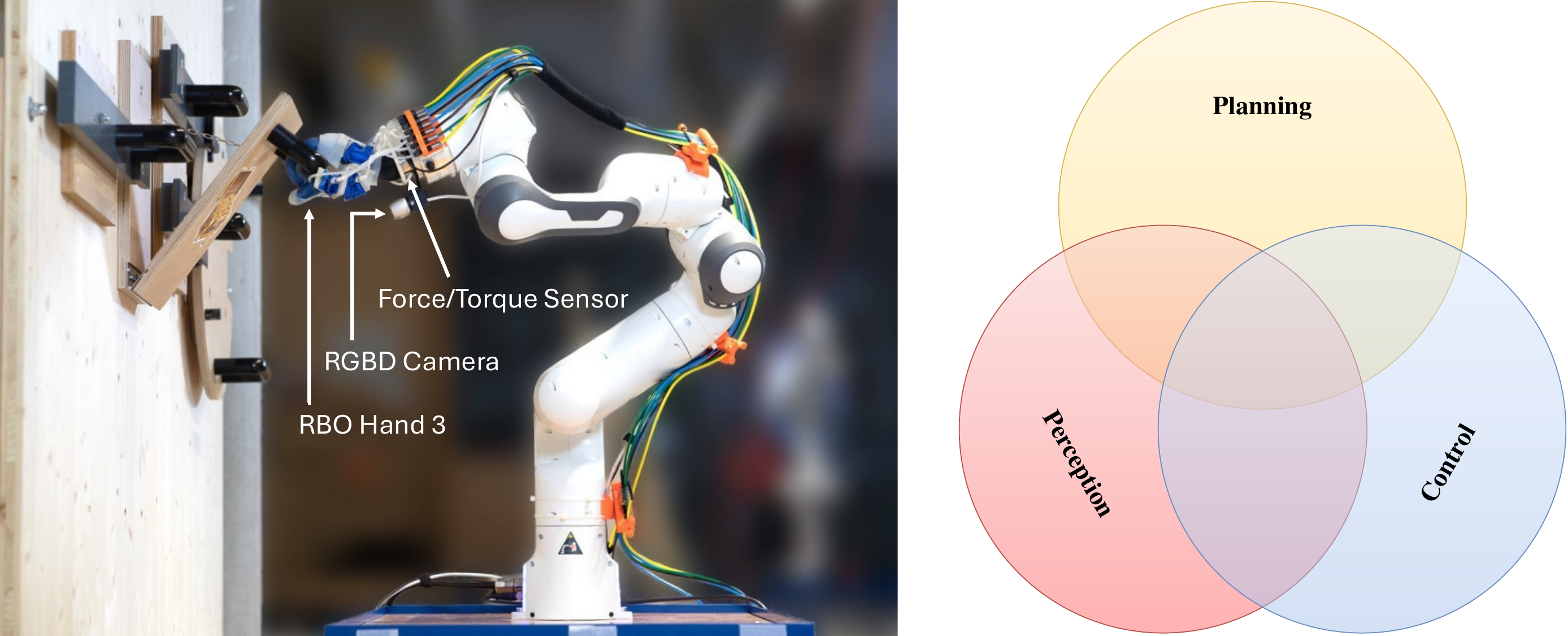}
	\vspace{0.5em}
		\captionof{figure}{Our robotic system solving a mechanical puzzle called lockbox (left). Our novel lockbox serves as a challenging testbed for evaluating the robustness of our system. Our hardware platform consists of a Franka Emika Panda arm equipped with an RBO Hand 3 end-effector~\cite{puhlmann2022rbo}, an RGB-D camera, and a Force/Torque sensor. The three fundamental components of our system are depicted as circles: \textit{Perception}, \textit{Control}, and \textit{Planning} (right). The four intersection regions of these components depict new behaviors emerging from actively interconnecting these components: a design principle inspired from biological agents. These active interconnections lead to an increased robustness in solving the lockbox as will be discussed in this paper.}
		\label{fig:titleFig1}
\end{figure*}

Robustness refers to the ability of a system to maintain performance under significant and unanticipated variability in the environment. This property is very desirable for robotic systems. However, outside of highly controlled environments, robotic systems rarely exhibit this property~\cite{krotkov2018darpa}. There is currently no systematic understanding of how robustness can be built into robotic systems. In contrast, nearly all biological systems exhibit significant levels of robustness. We suggest that this discrepancy is due to a lack of understanding of the factors that contribute to robustness in robotic systems. 

There are two main approaches to building robustness into robotic systems. The first one---let us call it system engineering---is motivated by best practices from engineering, and in particular software engineering. This school of system building is based on the assumption that robustness results from the encapsulation of complexity within the components, followed by composing the components into a system via simple interfaces~\cite{meyer1997object, martin2000design, booch2008object}. While this works well in the context of software, it does not work well for robotic systems that, unlike software, do not face well-defined inputs but instead must confront the unpredictability of the real world. System engineering advocates for a high degree of modularity, which is, as we will see, the opposite of how biological systems are ``engineered.''

The second, more recent approach to system building can be termed end-to-end or data-driven learning. At the moment of writing this paper, no substantial real-world robotic systems have been produced exclusively based on this approach. Most end-to-end learning systems also contain parts that were engineered in the sense above. Nevertheless, the promise is that these systems will one day be robust, i.e., generalize to out-of-distribution environmental conditions. The architectural pattern presented in this paper may shed some light onto why and how end-to-end learning might be able to achieve such generalization (more in Section~\ref{sec:relatedwork:part2}).

Both of these approaches have produced impressive robotic systems~\cite{schwarz2017nimbro, correll2016analysis, eppner2018four, rt2}, but neither has produced a systematic understanding of how to build systems for robustness.

Biological systems, much like robotic systems, are composed of individual components. In contrast to engineered robotic systems, however, biological systems utilize complex, versatile, and redundant connections between components, rather than simple interfaces. This means that the behavior of the whole system results not only from the individual components, but also---to a significant degree---from the interconnection between them. We will delve deeper into these biological design patterns in Section~\ref{sec:bio_interfaces}. We then apply the principle to the design of a complex, real-world robotic system, as shown in Figure~\ref{fig:titleFig1} and evaluate its performance in detail (Section~\ref{sec:experiment}). We finally provide possible explanations for why this design approach enhances robustness and some interesting future directions for this design principle (Section~\ref{sec:discussion}). Our extensive empirical evaluation of the resulting system confirms that this bio-inspired approach is indeed helpful for achieving robustness in robotic systems.

Our results indicate that we must rethink how we build complex robotic systems for robustness. The proposed principle gleaned from biological systems is probably just one of many such patterns. Of course, we only demonstrate the benefit of this principle in a single system. But we believe that our community should perform a systematic investigation of this and other architectural patterns for robustness. We therefore believe that the robotics community would benefit from a focused examination of the factors that contribute to robustness in robotic (and biological) systems.

\section{Inspiration From Biological Systems}
\label{sec:bio_interfaces}

While the robotics community lacks a systematic understanding of how to build robustness into robotic systems, biological systems exhibit a remarkable ability to adapt to various environmental circumstances~\cite{whitacre2012biological}. Giving this discrepancy, it does not seem far-fetched to attempt to learn something about robustness from biology. In this section, we examine several examples of striking biological robustness and attempt to extract a common pattern from them. This pattern then provides a specific hypothesis about how to achieve robustness in robotic systems. The rest of the paper is dedicated to our attempt to gather empirical evidence to assess the validity of the hypothesis.

To facilitate reading this section, we will start with the key message: Our hypothesis is that robustness in complex systems is produced by connecting the components of the system actively and in task-specific ways. By \textit{active} we mean that the connection does not simply pass information from one component to the other. Instead, an active interconnection considers the current state of the two components it connects and adjusts the information that is passed accordingly. And by \textit{task-specific} we mean that the way the information is actively adopted can depend on inductive biases. This can be either built into the system or derived in the interconnection from data. To summarize, our hypothesis is that robustness derives from \textit{active interconnections} among system components.

Let us now analyze some biological systems to see how this hypothesis is biologically inspired and supported. 

\subsection{Alternative Splicing in Protein Synthesis}
\label{subsec: alternative splicing}

In the cellular production of proteins, a process known as alternative splicing demonstrates \textit{active interconnections} at the level of genes~\cite{chen2009mechanisms}.

Alternative splicing is central to the versatility of the cellular protein production process in all living things. This process reads off genes to translate the genetic information into a protein. But rather than having a dedicated gene in the genetic code for each protein, the DNA encodes reusable building blocks. Alternative splicing assembles building blocks into larger pieces of genetic information from which the proteins are produced.

The input to protein production is a string, a one-dimensional sequence, the genetic information. The output is a protein, a three-dimensional, molecular structure. The genetic information, in effect, encodes this three-dimensional structure. It is important to note that the identical one-dimensional genetic building block, when pieced together with other building blocks, produces \textit{different} three-dimensional shapes in the complete protein. This variability results from the context a building block is placed in, in other words, from the other building blocks it gets connected to. (If Lego blocks had this property, the shape of the entire assembly would change with every added block.)

We now must consider the fact that already small variations in the genetic code can lead to the misfolding of proteins, when the protein does not assume the correct \mbox{3-D} shape for fulfilling its biological function, often leading to the death of the biological entity. How is it then possible that these building blocks can be assembled in many different ways by alternative splicing and still reliably lead to the biologically relevant structure of the protein?

The process of evolution has selected for building blocks that can cooperate in many different ways to produce biologically active proteins. Being put into a specific context, the parts of the protein corresponding to the building block exchange ``information'' during the folding process that leads to biological function. Building blocks must be able to adjust their ``behavior'' to their ``environment'' while still delivering a biologically functioning protein. The behavior of the protein (folding) is adjusted robustly by information exchange among the building blocks via physical forces that shape the folding process.

The robustness of protein folding from genetic material is based on versatile components (genetic building blocks) that exchange information (via physical force fields) to adjust their behavior (folding) to ensure a successful outcome (a biologically active protein).
The robustness of protein folding from genetic material is based on reusable components (genetic building blocks) that exchange information (via physical force fields) to adjust their behavior (folding) to ensure a successful outcome (a biologically active protein).

\subsection{Communication Between Cells}
\label{subsec: gap junctions}

In biological organisms, groups of cells cooperate to perform collective functions, for example in organs. While each cell can be viewed as a separate unit, their collective also operate as a unit, albeit at a different level of abstraction. This new level of abstraction requires information exchange between the cells for coordination.

Information exchange between cells is implemented in so-called gap junctions. These junctions facilitate versatile communication across cell membranes. It seems plausible that the variety of environmental conditions that cells are able to respond to (changes in temperature or pH level, invasion by pathogens, growth, differentiation, etc.) also necessitates equally diverse and tailored communication among the cells. This diversity in communication is indeed what we observe in cells~\cite{vaney2000gap}.

In this second example, we again encounter versatile information exchange among components (cells), actively adjusted to the context, leading to robustness, i.e., the ability to maintain performance under environmental variation.

\subsection{An Eye on the Back}
\label{subsec: tadpole}

Another instance of this pattern of active interconnections between components is particularly intriguing. When a tadpole eye is surgically attached to the tail of another tadpole, the tissue exhibits remarkable adaptability, allowing the tadpole to assimilate information from the new eye~\cite{ston2013ectopic}.

After the surgical attachment of the new eye, the tadpole's tissue responds dynamically to the unforeseen sensory input. It grows of a connection between the optic nerve of the eye and the tadpole's nervous system. Even more surprising, the tadpole is able to adapt its behavior based on light detected by the newly integrated eye. To put this into context, a tadpole is larval stage of amphibian life, when the organism is still being formed. This example shows, that these formation processes are highly robust to Frakensteinian tinkering. 

This example shows that the components of the tadpole's body are ready to interact with evolutionarily unplanned components in novel ways. Similar to the example on alternative splicing, this provides evidence that components are set up to exchange information with other components in surprising ways, enabling a meaningful behavioral response that contribute the robustness of the developmental process and the resulting biological system. 

\subsection{Transplanting Organs}
\label{subsec: transplants}

In transplantation medicine, we see another biological example that illustrates the complex interconnections between biological components. When intestines are transplanted, the incorporation of the liver in the transplant seems to decrease the likelihood of rejection of the implanted organs~\cite{calne1969induction, abrol2019revisiting}. This represents a broader trend wherein including a less immunogenic organ can enhance the long-term viability of a transplant involving a more immunogenic organ. This further underscores that advantageous interactions among organs extend beyond their immediate physiological functions. The robustness of the ensemble of organs seems to depend on more intricate interdependencies than our knowledge of the physiological functions would imply.

\subsection{A Lesson From Biology}
\label{subsec: AI lesson}

These examples showcase that components in biological systems exchange information actively, adjusting the information to the specific context. This active information exchange seems to be a precondition for robustness. As a result of engaging in active information exchange among components, biological systems are able to maintain their performance under environmental variations. We believe that active interconnections among components are a key architectural principle of the robustness exhibited by biological systems. 

In this work, we want to find out if the hypothesized architectural principle of active interconnection, when transferred to robotics, also leads to robustness. We only draw inspiration from the biological examples we introduced, but will not attempt to imitate them.

\section{Analysis of Robotic Systems}
\label{sec:relatedwork}

The proposed biologically inspired design principle advocates for establishing active interconnections between system components to enhance robustness against environmental variations. As the interest in building robust robotic systems has been longstanding, in our discussion of related work, we analyze existing robotic systems to identify potential instances where this biological principle has been implemented and assess their effectiveness in achieving increased robustness.

\subsection{Systems with High-Degree Modularity}
\label{sec:relatedwork:part1}

Robotic systems are often designed by decomposing systems into distinct components that interact with each other through simple and well-defined interfaces. This strategy, which emphasizes a high degree of modularity---a principle valued in software engineering---offers several advantages. First, components with simple and well-defined interfaces are more easily replaceable, modifiable, and expandable~\cite{compUpg1, hernandez2017team}. Additionally, by adopting the software engineering concept of encapsulation, component interconnections are often specified in the early stages of system design. This enables parallel development as developers can focus on individual components without requiring a comprehensive understanding of the entire system~\cite{parallel1, parallel2}. Furthermore, strong modularization confines errors and failure modes to their corresponding components, preventing their propagation throughout the system~\cite{errorPro1, errorPro2, errorPro3}, thus simplifying the debugging process. 

These benefits have led to the widespread adoption of the high-degree modularity in developing reusable robotic frameworks~\cite{quigley2009ros} and libraries~\cite{chitta2012moveit, rusu20113d}.
This design principle has also demonstrated its effectiveness at building well-performing robotic systems in large-scale robotics challenges, such as the DARPA Robotics Challenge (DRC)~\cite{righetti2014autonomous} and the Amazon Picking Challenge (APC)~\cite{hernandez2017team, matsumoto2020end}.

Systems resulting from high-degree modularity are typically characterized by isolated components and limited interconnections. Although the success of highly modular systems seems to contradict our hypothesized design principle, it's crucial to recognize that high-degree modularity primarily enhances robustness from a software engineering perspective. Software products often have well-defined requirements, and the main challenge is minimizing human errors to maximize economic returns. High-degree modularity addresses these challenges by optimizing the development process to improve robustness against human errors. However, while this type of robustness is desirable for software products, it fundamentally differs from the robustness required by robots to maintain system performance under significant, unpredictable environmental variations. In this work, we explore whether we can achieve this latter type of robustness by establishing diverse and active interconnections among system components.

\subsection{Systems With Enhanced Component Interconnections}
\label{sec:relatedwork:part2}

In contrast to highly modular systems, which limit component interconnections according to the software principle of encapsulation, we have observed works that generate robust robotic behaviors by enhancing the interactions among components. For instance, prior works have explored interconnections between perception and control to develop reactive controllers, such as tactile and visual servoing controllers~\cite{servo1, servo2, servo3, servo4, schwarz2017nimbro}, that demonstrate robustness to local environmental changes and uncertainties in perception and actuation. 
However, reactive controllers are inherently limited by their susceptibility to local minima when accomplishing long horizon tasks, especially in complex manipulation tasks, where these controllers are particularly prone to perceptual aliasing. Reactive planning addresses this limitation by advocating additional active interconnections between planning, perception, and control. 
For example, action outcomes can inform planning to select appropriate controllers that react properly to perceived environmental changes~\cite{mosaic1, mosaic2}. Furthermore, integrating perception and control into planning enables systems to use locally gained information to adjust global plans, resulting in more robust behaviors~\cite{majumdar2017funnel, kappler2018real, baum2022world, toussaint2022sequence}.

Interactive Perception~\cite{bohg2017interactive} is another paradigm supporting the idea of active interconnections leading to robust robotic behaviors. It exploits the regular relationship between actions and the corresponding sensory output to simplify manipulation through exploiting environmental constraints~\cite{eppner2015exploitation, pisaHand, ece1, ece2}, facilitate failure detection~\cite{morrow1997manipulation, baum17}, perform system identification~\cite{zhong2023robot} and enhance the performance of exploration~\cite{ge1, ge3}.

While the presented examples demonstrate the potential of active interconnectivity to generate robust robotic behaviors in response to environmental changes, no study has yet analyzed the correlation between active interconnection and system-level robustness. To address this gap, we apply this design principle to build a system from scratch, focusing on maximizing advantageous active interconnections among components. In this way, we aim to gain a more comprehensive understanding of the relationship between this design principle and robustness against environmental variations.

\subsection{End-to-End Learning Systems}
\label{sec:relatedwork:part3}

The concept of active interconnections extends beyond engineering-based robotic systems, appearing prominently in the growing trend of end-to-end learning systems~\cite{trendInRobotLearning}. These systems aim to optimize components jointly within a unified framework, effectively blurring the boundaries between them. For instance, many studies demonstrate that robotic behaviors can be acquired by jointly training perception and control components using human demonstration data~\cite{rt1, act, diffusion}.

Building on this approach, research has shown that merging the perception and planning components~\cite{palem} can enhance planning success and improve robustness against environmental variations. Furthermore, jointly training perception, control, and planning as a single system~\cite{rt2} has been found to result in more robust robotic behaviors in dynamic environments.

From the perspective of active interconnections, end-to-end learning systems achieve a more comprehensive form of interconnectivity without information loss compared to engineered systems. However, recent benchmarks indicate that current end-to-end learning systems struggle with complex long-horizon manipulation tasks in the real world. For instance, \cite{heo2023furniturebench} reports that despite 1,000 demonstrations and minimal environmental changes, the end-to-end systems are still unable to solve real-world long-horizon manipulation tasks. Similar findings are reported in~\cite{lee2021ikea, pumacay2024colosseum, luo2024fmb}. These results indicate that current learning algorithms still face challenges in distinguishing task-relevant information from instance-specific and environmental details, thereby requiring proper inductive biases to achieve robustness, i.e., generalization to out-of-distribution environmental conditions~\cite{zhang2024inductive}. Nevertheless, our arguments supporting active interconnections suggest that end-to-end learning systems have the potential to automatically identify task-relevant inductive biases, leading to highly robust robotic systems in the future.

\section{The Lockbox Problem}
\label{sec:lockboxenvironment}

To evaluate our biologically inspired design principle for enhancing a robotic system's robustness, we need a complex task. Complex tasks necessitate high degree of coordination between system components, making them suitable for studying and evaluating active interconnections. Since we aim to evaluate the system’s robustness against environmental variations, the task setting should be easy to modify. With these criteria in mind, we propose the lockbox problem as a suitable benchmark.

\subsection{Lockboxes as an Ideal Testbed for Assessing Robustness}
\label{sec:motviation}

To achieve a level of robustness comparable to that of biological agents, it is logical to use a testbed that is also used to evaluate cognitive processes in these agents. Lockboxes are mechanical puzzles with interlocking movable joints, require sequential manipulation to reach a specific goal state. These puzzles are ideal for this purpose due to their high cognitive demands, which have been widely used for studying intelligence in various species, including cockatoos~\cite{auersperg2011flexibility, auersperg2013explorative}, keas~\cite{huber2001social, miyata2011keas}, cats~\cite{chance1999thorndike}, mice~\cite{lang2023challenges}, elephants~\cite{jacobson2022persistence}, raccoons~\cite{Daniels2019BehavioralFO} and primates~\cite{carvalho2008chaines, TECWYN2013174, canteloup2020wild}. 

Solving a lockbox requires long-term planning and intricate interconnections among multiple system components. For example, the puzzle often exceeds an agent's perceptual range, necessitating active viewpoint adjustments---a crucial interplay between perception and control. Moreover, understanding the lockbox’s structural intricacies, such as the number and arrangement of joints, is essential for effective planning. These challenges underscore the need for robust communication and coordination among system components, making the lockbox problem a compelling benchmark for examining the impact of active interconnections on overall system robustness.

Lockboxes also provide a versatile platform for exploring the relationship between complex component interconnections and robust adaptation to environmental changes. For example by varying joint types and interlocking dependencies, we can systematically assess a robotic system’s ability to handle diverse task conditions. Different mechanical joint types require distinct manipulation strategies, while altering the joint arrangement changes the problem’s structure and difficulty. This flexibility allows for a comprehensive evaluation of system robustness across a broad spectrum of environmental challenges.

\subsection{Our Lockbox}
\label{sec:lockboxproblem}


Building on previous works~\cite{baum2017opening, verghese2023using, ota2023h, liu2023busybot, abbatematteosensorized}, we propose our own version of the lockbox, which introduces key distinctions to increase the problem's complexity. Similar to related studies, our lockbox incorporates binary joint states, meaning that each joint can only be positioned at either end of its motion range. The lockbox is considered solved when a pre-specified joint, which we refer to as \textit{target joint}, is moved to the end of its motion range.

Two key characteristics distinguish our lockbox problem from previous work. First, the robot must autonomously determine task-relevant properties of the lockbox, including joint configuration, joint count, grasp poses, and manipulation policies. This is a prerequisite for assessing the robustness of robotic behaviors in the real world. 

Second, our lockbox contains three types of joint interdependencies with the last one significantly increasing the task complexity:

\begin{itemize}
	\item \textbf{One-to-One}: one joint locks another single joint
	\item \textbf{Many-to-One}: multiple joints lock one single joint
	\item \textbf{Bistable-Locking}: one joint locks multiple joints at two different states
\end{itemize}

\begin{figure}[ht]
	\vspace{0.5em}
	\centering
	\begin{subfigure}[b]{0.11\textwidth}
		\centering
		\includegraphics[width=\textwidth]{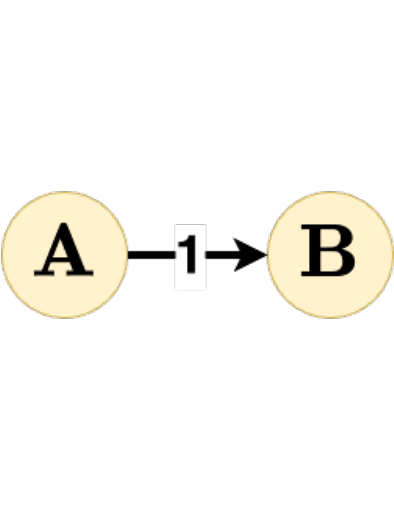}
		\caption{\tparbox{2.5cm}{One-to-One Dependency}}
		\label{fig:one2one}
	\end{subfigure}
	\hfill
	\begin{subfigure}[b]{0.11\textwidth}
		\centering
		\includegraphics[width=\textwidth]{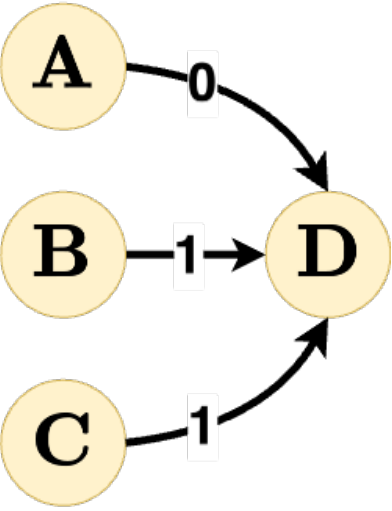}
		\caption{\tparbox{2.5cm}{Many-to-One Dependency}}
		\label{fig:many2one}
	\end{subfigure}
	\hfill
	\begin{subfigure}[b]{0.11\textwidth}
		\centering
		\includegraphics[width=\textwidth]{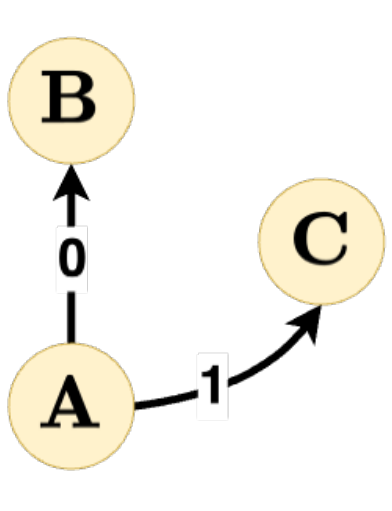}
		\caption{\tparbox{2.5cm}{Bistable-Locking Dependency}}
		\label{fig:bi_stable}
	\end{subfigure}
	\caption{Joint interdependencies as \textit{Directed Acyclic Graphs} (DAGs): The One-to-One example (Figure~\ref{fig:one2one}) depicts joint $\mathbf{B}$ depending on joint $\mathbf{A}$ being in state 1. The Many-to-One example (Figure~\ref{fig:many2one}) shows joint $\mathbf{D}$ requiring joints $\mathbf{A}$, $\mathbf{B}$, and $\mathbf{C}$ in the respective states 0, 1, and 1 to unlock. Joint $\mathbf{A}$, in Figure~\ref{fig:bi_stable} is a bistable-locking joint, which locks joints $\mathbf{B}$ and $\mathbf{C}$ in two different states.}
	\vspace{-1em}
	\label{fig:dependencies}
\end{figure}

These three joint interdependencies can be represented as \textit{Directed Acyclic Graphs} (DAGs) as visualized in Figure~\ref{fig:dependencies}. In these graphs, nodes depict the lockbox’s joints, while directed edges point from locking joints to joints they lock. The edge value, either 0 or 1, indicates the state that the locking joint must be in for the connected joint to be unlocked. A joint is unlocked only if all its locking joints have state values matching to the corresponding edge values.

It is important to note that introducing Bistable-Locking (BL) joints significantly increases the number of possible manipulation steps required to solve the lockbox. This is also showcased in a study where the number of steps humans require when solving lockboxes significantly increases as the numbers of BL joints increases~\cite{zenkri_bolenz_pachur_brock_2024}.

\begin{figure}[H]
	\centering
	\includegraphics[width=1\linewidth]{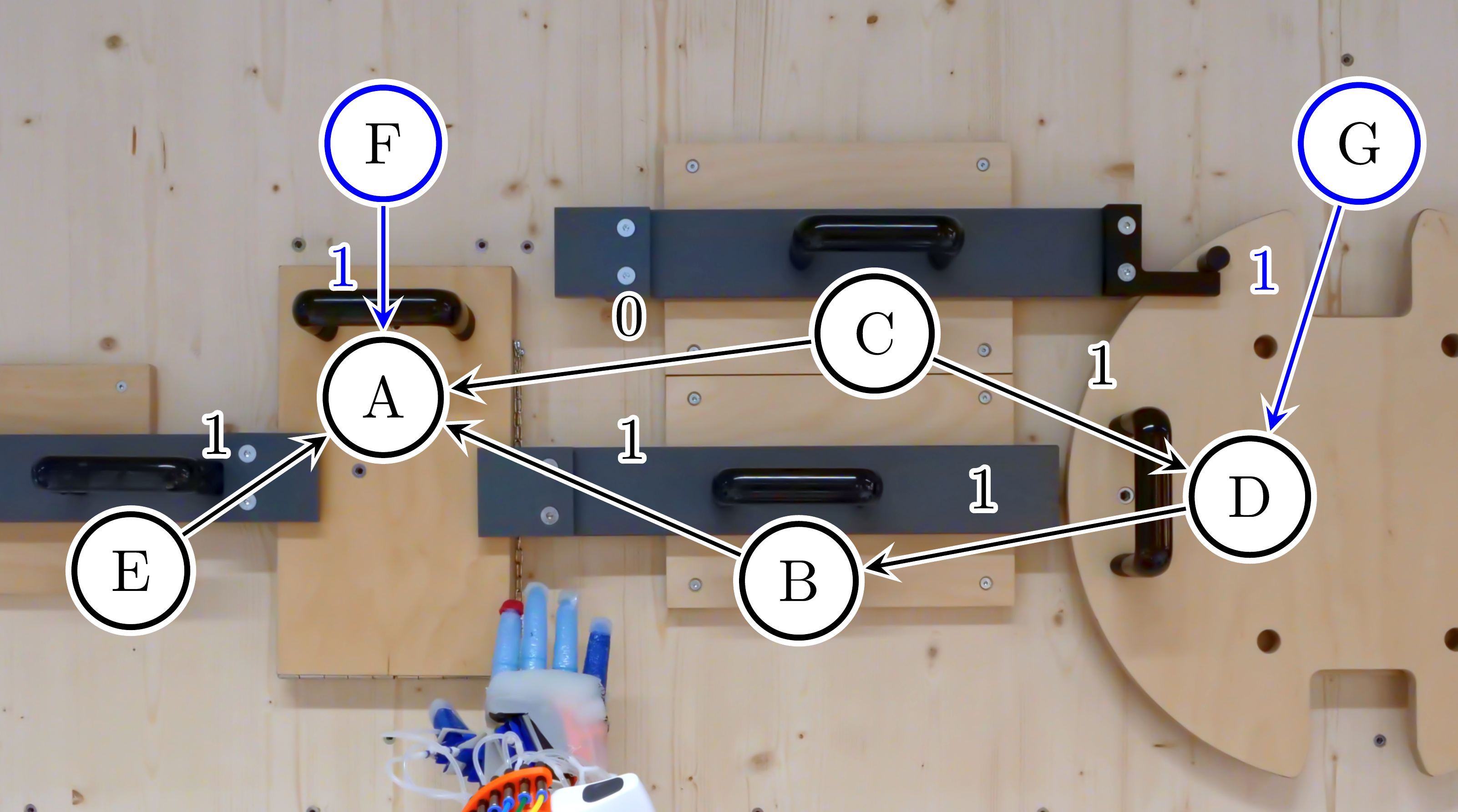}
	\caption{Our physical Lockbox overlayed by the joint-interdependencies graph. The \textcolor{blue}{blue} marked nodes depict the additional fictive joints used in the simulation. The directed acyclic graph visualizes the interdependency structure.}
	\label{fig:physicalLockbox}
\end{figure}

We now introduce our physical lockbox. As shown in Figure~\ref{fig:physicalLockbox}, our physical lockbox consists of three prismatic joints (joints $\mathbf{B}$, $\mathbf{C}$, and $\mathbf{E}$) and two revolute joints (joints $\mathbf{A}$ and $\mathbf{D}$) mounted on a wooden wall. For our simulation experiments, we extend the physical lockbox with two virtual prismatic joints, marked in blue (joints $\mathbf{F}$ and $\mathbf{G}$). The interlocking dependency is illustrated as a DAG. Specifically, joints $\mathbf{B}$ and $\mathbf{D}$ exemplify a \textit{one-to-one} dependency, where the locking state of joint $\mathbf{B}$ depends solely on the state of joint $\mathbf{D}$. Joints $\mathbf{B}$, $\mathbf{F}$, and $\mathbf{E}$ exhibit a \textit{many-to-one} dependency on joint $\mathbf{A}$. Finally, joint $\mathbf{C}$ functions as a \textit{bistable-locking} joint, restricting the manipulation of either $\mathbf{A}$ or $\mathbf{D}$ depending on its state (0 for $\mathbf{D}$, 1 for $\mathbf{A}$). As we explained earlier, introducing bistable-locking dependencies to the other types of dependencies greatly increases the lockbox's complexity.

\section{Example System with Active Interconnections}
\label{sec:Systems}

This section outlines our approach to applying the design principle of active interconnections to construct a robotic system capable of solving lockboxes, as shown in Fig~\ref{fig:titleFig1}. We implement active interconnections as intermediary entities that dynamically modulate information flow between system components in task-specific ways, generating additional information or behaviors that otherwise not available. Notably, distinct interconnections between the same components can produce qualitatively different behaviors, emphasizing the compositional nature of this design principle.

We build our robotic system around three fundamental components \emph{perception}, \emph{control}, and \emph{planning}. Each component plays a distinct role in addressing the lockbox-solving challenge: perceiving action affordances (perception), physically manipulating mechanical joints (control), and reasoning about the underlying interlocking dependencies (planning). We start by describing these three fundamental components and then introduce the active interconnections we have established among them and explain how they contribute to generating behaviors that enhance the system’s robustness to environmental changes.

\subsection{Components}
\label{sec:system:components}

\subsubsection{Perception}
\label{sec:perception}

In the context of a lockbox manipulation task, a perception component needs to fulfill at least two requirements:~1) Handle detection and tracking and~2) Grasp pose generation. For the first requirement, since our lockbox consists of mechanical joints with standard door handles, we adopt an off-the-shelf object detector based on Faster R-CNN~\cite{Ren2015fasterrcnn} and fine-tune it on a publicly available door-handle dataset for the handle detection~\cite{arduengo2021handledataset}. We employ the Hungarian algorithm with a Kalman filter to track the detected handles over time~\cite{Hamuda2018ImprovedImageProcessing}. Tracking is required as the robot has to actively change its viewpoint in order to recognize all handles of the lockbox.

Next, we explain the process of computing a grasp pose for each detected handle. We assume that handles are mounted parallel to the plane of the lockbox wall. We define a handle pose by selecting the longest side of the handle as the y-axis and aligning the z-axis with the plane's normal direction. We then apply \textit{Principal Component Analysis} (PCA) on the detected bounding box to estimate the orientation of the handle and use depth measurements to determine its position relative to the robot base frame. After obtaining a handle pose in the robot base frame, we compute a relative transformation from the handle pose to a grasp pose for the grasping behavior. We learn this relative transformation via Programming by Demonstration~\cite{billard2008survey}. Concretely, given a handle pose H in the robot base frame B namely $T_\mathrm{BH}$ in SE(3), we manually move the end-effector, i.e., RBO Hand 3 to a desired grasp pose G, and record the pose $T_\mathrm{BG}$. The relative transformation can be computed by $T_\mathrm{HG} = T^{-1}_\mathrm{BH} \cdot T_\mathrm{BG}$. For a newly detected handle pose $T_\mathrm{BH}$, we can compute the grasp pose with this relative transformation as $T_\mathrm{BG} = T_\mathrm{BH} \cdot T_\mathrm{HG}$.

\subsubsection{Control}
\label{sec:control}

The control component is responsible for grasping and interacting with joints. We use a Jacobian-transposed-based Cartesian impedance controller to move the robot towards a desired end-effector pose~\cite{khatib1987unified}. We assume that linear interpolation can be used to generate feasible trajectories in Cartesian space without requiring advanced motion planning algorithms.

We now describe two important manipulation behaviors for solving lockboxes. First, to grasp a joint, we control the arm to the desired grasp pose in SE(3) and close the robot hand. Once the grasp is established, the robot wiggles the joint by sequentially executing six straight-line movements along the axis directions within the end-effector coordinate system. If the maximum observed movement exceeds a predefined threshold, the manipulated joint is identified as movable. An innovative aspect of our system is its ability to autonomously manipulate various types of joints. This is achieved by establishing an active interconnection between perception and control, as explained later in Section~\ref{sec:ConstraintFollowing}.

\subsubsection{Planning}                                                  \label{sec:planning}

A planning component is needed to perform high-level reasoning to efficiently solve a lockbox. It is important to mention that prior works on lockboxes assume that each joint is constrained by only one other joint or a joint can lock other joints in only one state~\cite{baum2017opening, verghese2023using}. Therefore, their planning methods cannot be applied to our lockbox, which includes three types of joint interdependencies: one-to-one, many-to-one, and bistable-locking, as explained in Section~\ref{sec:lockboxenvironment}. We have to design a planning component to solve our lockboxes with increased complexity. To do so, we drew inspiration from human exploratory behaviors. We conducted an experiment where human participants solved lockboxes with various scales and interlocking dependencies in simulations. By analyzing how participants solved the lockboxes through interactions, we formulated a heuristic-based solver and incorporated it into our planning component.

\begin{algorithm}[htp]
	\SetAlgoLined
	\LinesNumbered
	\KwResult{Solve the lockbox puzzle}
	\caption{Human-Inspired Heuristic for Solving Lockboxes}
	\label{alg:heuristic}
	$\text{movable}_j \gets \text{false},~\forall j \in \mathbb{J}$\;
	\ForEach{$j \in \mathbb{J}$}{
		$\text{moved} \gets$ manipulate($j$)\;
		$\text{movable}_j \gets \text{true}$ \textbf{if} moved \textbf{else} $\text{false}$\;
	}
	$\mathbb{J}_{\text{free}}$, $\mathbb{J}_{\text{locked}} \gets$ split($\mathbb{J}$, movable)\;
	
	\While{$\lnot~{\mathrm{movable}}_{j_{\text{target}}}$}{
		$\mathrm{C} \gets \text{combinations}^*(\mathbb{J}_{\text{free}})$\;
		\ForEach{$c \in \mathrm{C}$}{
			\ForEach{$j \in c$}{
				$\text{moved} \gets$ manipulate($j$)\;
				\If{$\lnot~\mathrm{moved}$}{
					$\text{movable}_j \gets \text{false}$\;
					$\mathbb{J}_{\text{free}}$, $\mathbb{J}_{\text{locked}} \gets$ split($\mathbb{J}$, movable)\;
					\textbf{break} \tcp*{Break out of the for loops}
				}
			}
			\ForEach{$j \in \mathbb{J}_{\text{locked}}^*$}{
				$\text{moved} \gets$ manipulate($j$)\;
				\If{$\mathrm{moved}$}{
					$\text{movable}_j \gets \text{true}$\;
					$\mathbb{J}_{\text{free}}$, $\mathbb{J}_{\text{locked}} \gets$ split($\mathbb{J}$, movable)\;
					\textbf{break} \tcp*{Break out of the for loops}
				}
			}
		}
	}
\end{algorithm}

This heuristic-based solver, described in Algorithm~\ref{alg:heuristic}, operates as follows: It starts by assessing the locking state (movable or locked) of every joint $j$ from the set of detected joints $\mathbb{J}$ (lines 1 to 5). Joints that can be moved are added to the set of free joints $\mathbb{J}_{\text{free}}$, and the locked ones to the set of locked joints $\mathbb{J}_{\text{locked}}$ (line 6). Now the algorithm iterates over the next two steps until the target joint $j_{\text{target}}$ becomes movable, i.e. the lockbox is solved. The first step is to realize a state-combination $c$ from the set of all possible combinations $\mathrm{C}$, that can be realized with the free joints (lines 10 to 17). Second, every joint from the set of locked joints is tried (lines 18 to 25). During these two steps, if a locked joint moves (second step) or a joint assumed movable becomes locked (first step), the current attempt is abandoned. This triggers a restart of the iterative process with the updated combinations set $\mathrm{C}$, taking the changes to $\mathbb{J}_{\text{free}}$ into consideration (line 8).

The sequence of free joint combinations and the order of attempting locked joints are crucial for efficiently solving a lockbox.  In the next subsection, we will explain how to achieve efficient solving behaviors by leveraging active interconnections between perception, control, and planning to create an attention mechanism. This attention mechanism will guide the heuristic-based solver in the elements marked with~* (line 8 and 18).

\subsection{Active Interconnections}
\label{sec:RichComponentInteractions}

\begin{figure}[H]
	\centering
	\includegraphics[width=0.8\linewidth]{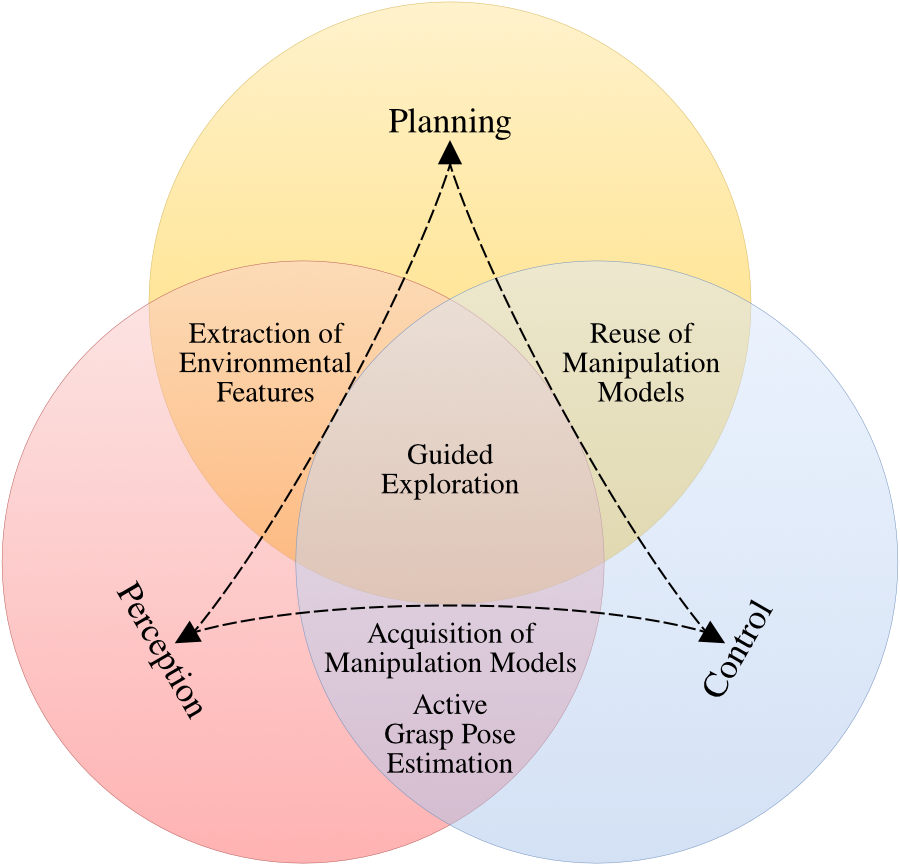}
	\caption{Illustration of our robotic system with different active interconnections among components. The fundamental components—perception, control, and planning—are depicted as circles, with overlapping regions showcasing new behaviors resulting from the active interconnections.}
	\label{fig:AI}
\end{figure}

Having introduced the three fundamental components of our system, we now explain how to establish active interconnections among them, as illustrated in Figure~\ref{fig:AI}, and elaborate on how the emerging behaviors promote robustness against environmental variations.

\subsubsection{Acquisition of Manipulation Models}
\label{sec:ConstraintFollowing}

To solve the lockbox, the robot must first manipulate the mechanical joints constituting the lockbox. This manipulation requires a model of the kinematic structure of those mechanical joints. Rather than first acquiring such a model to then enable control to operate the joint, we create an active interconnection between perception and control for the model building. In this active interconnection, information about the mechanical joint (obtained initially from random motion) is used to enable effective manipulation of the joint. This, in turn, creates more motion in the mechanism, revealing more information about the joint. The active interconnection produces a cycle of mutual information exchange that produces robust motion and produces an accurate model of the mechanical joint, at the same time. By actively interconnecting perception and control, we have reduced the system’s dependency on models and made it robust to model inaccuracies. We have improved the robustness of the system by adding an active interconnection.

The active interconnection is implemented as follows: It begins by using the perceived end-effector displacement, $\Delta p \in \mathbb{R}^3$, obtained initially from the robot’s wiggling behavior (see Section~\ref{sec:control}), to compute the admissible motion direction $d=\frac{\Delta p}{|\Delta p|}$ of the manipulated joint. The control component then directs the end-effector along this computed direction to reveal more end-effector motion. This, in turn, facilitates estimating the joint's admissible motion direction, thus closing the feedback loop. This active interconnection simultaneously estimates and guides the robot toward the joint's admissible motion direction, enabling the robot to manipulate various mechanical joints with 1 DoF.

Note that applying this active interconnection with a soft end-effector is not straightforward. The deformations of the soft end-effector during forceful interactions can lead to noisy end-effector displacement observations, resulting in poorly estimated admissible motion directions. This inaccuracy may cause the robot to lose contact with the mechanical joint's handle, leading to manipulation failures. To address this issue, we introduce an additional active interconnection between perception and control to regulate the wrench during manipulation, thereby minimizing the deformation of the soft end-effector.

This additional active interconnection works as follows: Given a desired pose D in the base frame B namely $T_\mathrm{BD}$, we first interpolate the goal in the exponential coordinate $\Delta\psi_\mathrm{t} \in se(3)$ with a twist limit $V_\mathrm{m} \in \mathbb{R}^6$ as:

\begin{equation}
	\Delta\psi_\mathrm{t} = \max\left( \min\left( \mathrm{log}(T^{*-1}_\mathrm{BE_{t-1}} \cdot T_\mathrm{BD}), \ V_\mathrm{m}\Delta t\right), \ -V_\mathrm{m}\Delta t \right),
\end{equation}

where $T_\mathrm{BE_{t-1}}$ is the interpolated pose in the timestamp $\mathrm{t-1}$ and $\Delta t$ is the control period. We then constraint $\Delta\psi_\mathrm{t}$ based on the difference between the observed wrench $F_\mathrm{obs} \in \mathbb{R}^6$ measured by the wrist-mounted Force/Torque sensor and a wrench limit $F_\mathrm{m} \in \mathbb{R}^6$:

\begin{align}
	\Delta F_\mathrm{t} &= F_\mathrm{m} - |F_{\mathrm{obs, t}}|, \\
	\Delta\psi^*_\mathrm{t} &= \left\{\begin{array}{ll}
		\tanh(|\Delta F_\mathrm{t}|) \cdot \Delta\psi_\mathrm{t},
		& \text{if } \Delta F_\mathrm{t} > 0, \\
		\mathrm{sign}\left(F_{\mathrm{obs, t}}\right) \cdot \tanh(|\Delta F_\mathrm{t}|) \cdot V_\mathrm{m}\Delta t,
		& \text{otherwise,}
	\end{array}\right.
	\label{eq:test}
\end{align}

where the $\tanh(\cdot)$ function is used as a smoothing factor. When $\Delta F_{t} < 0$ exceeds wrench limits, $\mathrm{sign(\cdot)}$ flips the interpolation direction to reduce the observed wrench $F_{\mathrm{obs, t}}$. After obtaining the wrench-limited relative pose $\Delta\psi^*_\mathrm{t} \in se(3)$, we can compute the desired end-effector pose $T^*_\mathrm{BE_{t}}~=~T^*_\mathrm{BE_{t-1}} \cdot~\exp(\Delta\psi^*_{t})$ and move the robot using an Cartesian impedance controller~\cite{khatib1987unified}.

\begin{figure}[htp]
	\centering
	\includegraphics[width=\columnwidth]{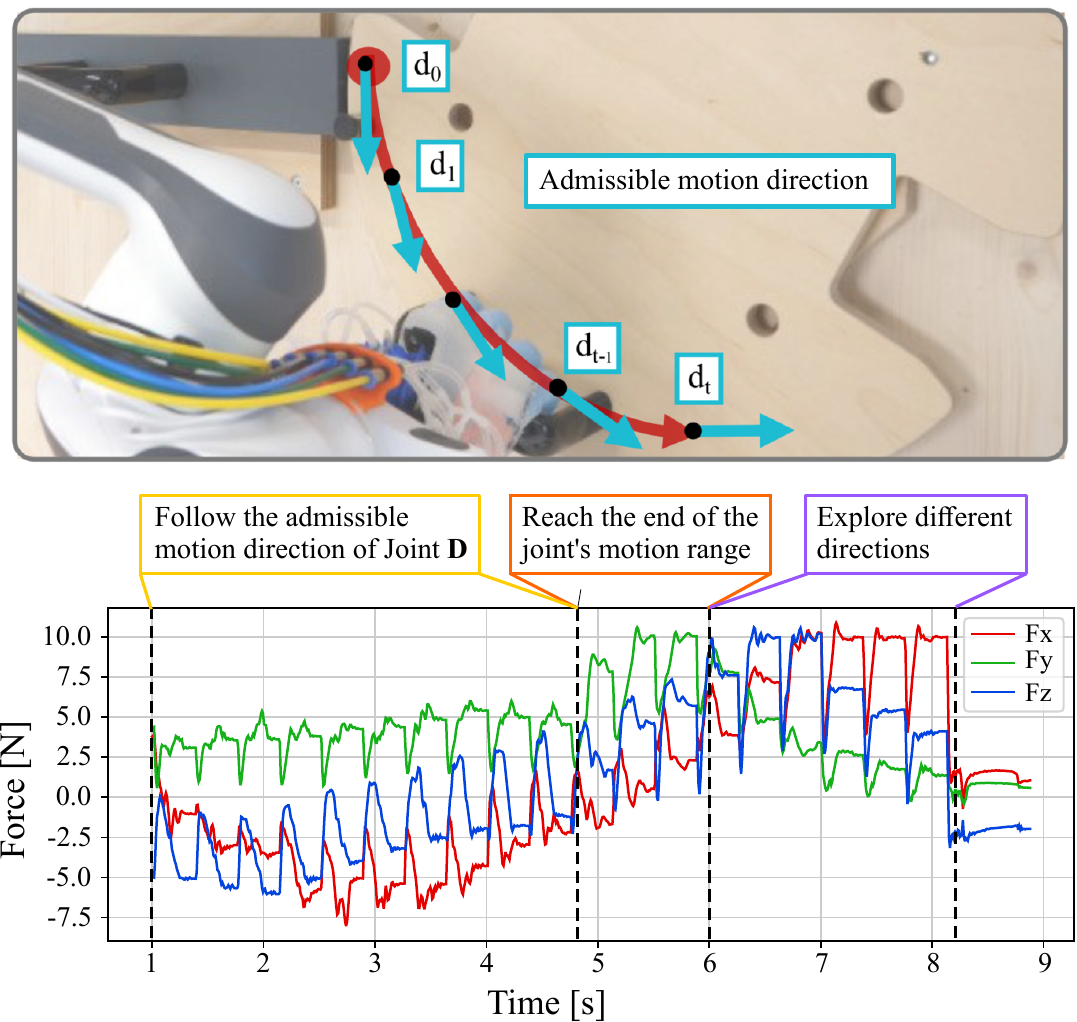}
	\caption{Autonomous manipulation of joint~\textbf{D} by simultaneous estimation and following of the admissible motion direction. The force measurements observed in the end-effector frame are illustrated below, depicting three distinct phases of manipulation. In the first phase (highlighted in yellow), the robot successfully follows the admissible motion direction of joint~\textbf{D}, aligned with the y-axis of the end-effector's frame. As the robot arrives at the motion limits of joint~\textbf{D}, it also reaches the predefined force limit of 10N in the y-axis direction (shown in orange). Subsequently, the robot begins exploring other directions (indicated in purple), leading to different force measurements. However, movement in these directions is undesirable, as they result from deformations of the soft end-effector rather than from joint~\textbf{D}, potentially causing issues like losing contact with the handle. By establishing an active interconnection, the robot efficiently regulates force within the predefined force limit (10N) during this exploration, ensuring successful manipulation with the soft end-effector.}
	\label{fig:constraintFollowingForce}
\end{figure}

The key idea of this active interconnection is to integrate the perceived wrench measurements into the trajectory generator that rapidly adjusts the equilibrium pose $T^*_\mathrm{BE_{t}}$, allowing the robot to manipulate joints within a predefined wrench threshold, thus limiting the undesired end-effector’s deformations. An example of this active interconnection in manipulating joint~\textbf{D} is visualized in Fig~\ref{fig:constraintFollowingForce}. This active interconnection is crucial for accurately estimating the admissible motion direction during manipulation with a soft end-effector, preventing manipulation failures such as loss of grasp, thus contributing to robust forceful manipulation behaviors.

\subsubsection{Extraction of Environmental Features}
\label{sec:ExtractionOfEnvironmentalFeastures}

The aforementioned active interconnections exploit the interdependence between sensory observation and action, enabling the robot to autonomously acquire manipulation models (i.e., end-effector trajectories), required for manipulating mechanical joints. In addition to manipulating the lockbox’s individual joints, the robot must also reason about their interlocking dependencies to solve the lockbox efficiently. In addition to assessing the binary state of each mechanical joint (locked or unlocked), the features of the mechanical joints, e.g., positions or kinematic joint types, could also be useful to improve the solving efficiency. For example, in a physical lockbox, nearby mechanical joints are more likely to influence each other than distant ones. Therefore, establishing an active interconnection that exploits the perceptual information about the environment to support the planning component can lead to more robust systems.

For our system, we consider two environmental features that could complement the planning component in solving lockboxes: the Cartesian positions of each joint handle, denoted as $\left(x, y, z\right) \in \mathbb{R}^3$, and its kinematic joint type $k \in \mathbb{R}$. We determine the handle positions during the handle detection process (Section~\ref{sec:perception}). For the kinematic joint type detection $k$, we apply PCA on the performed end-effector trajectory after a successful joint manipulation:
\[
k= 
\begin{dcases}
	1, \: \text{prismatic}, & \text{if } \lambda_{1} > (1 - \alpha) \cdot S\\
	-1, \: \text{revolute}, & \text{if } \lambda_{3} < \alpha < \lambda_{2}\\
	0,            & \text{otherwise,}
\end{dcases}
\]
where $\lambda_1, \lambda_2, \lambda_3 \in \mathbb{R}^3$ denote the first three eigenvalues of the end-effector trajectory sorted in descending order, $S = \sum\nolimits_{i=1}^{3}\lambda_{i}$ denotes the sum of all eigenvalues, and $\alpha$ is the threshold for distinguishing prismatic from revolute joints. A joint type of 0 is attributed to joints that have not been successfully manipulated yet, i.e. the end-effector trajectory is still unknown. Next, we explain how these environmental features can be used to inform the planning component by another active interconnection. 

\subsubsection{Guided Explorations}
\label{sec:GuidedExplorations}

We now introduce an active interconnection that extracts potential interlocking dependency patterns from the aforementioned environmental features. This active interconnection works as an attention system that prioritizes which joints to attempt next. Instead of relying on preprogrammed prioritization patterns—such as a mechanical joint is more likely to unlock closer joints than joints farther away—we designed this active interconnection with the ability to adapt its prioritization dynamically based on previous interaction experiences. This online adaptation capability eliminates the dependency on predefined patterns of interlocking dependency, allowing the system to adjust to a wide range of configurations beyond pre-specified changes.

The active interconnection operates as follows: it computes the absolute differences in environmental features between the last manipulated joint and the remaining joints in the lockbox, namely the changes in position along each axis $\left(|\Delta x|, |\Delta y|, |\Delta z|\right)$ and differences in joint type $\left(\Delta k\right)$. The active interconnection then processes these feature differences to calculate a manipulation-priority score for each joint. Joints with higher scores are more likely to be successfully manipulated and are therefore prioritized in the planning sequence. Importantly, this active interconnection is based on a ridge regression model~\cite{hoerl1970ridge}, which is lightweight and allows the model's weights to be updated online without pre-training. Specifically, we update the weights using the results of the last five trials, labeling a trial as 1 if the robot successfully operates a joint and -1 otherwise.

The contribution to the robustness of this active interconnection is grounded on the ability to identify and exploit the interlocking patterns in an online manner. As demonstrated in our later experiment (Figure~\ref{fig:onlineWeights}), this active interconnection can effectively adapt to various lockbox configurations by online learning these patterns from the interaction experience.

\subsubsection{Reuse of Manipulation Models}
\label{sec:ReuseOfManipulationModels}

The control component significantly benefits from knowing the joint states (0 or 1) provided by the planning component. When encountering a previously manipulated joint, the robot can reuse the acquired manipulation models i.e., end-effector trajectory to manipulate the joint, avoiding possible failures resulting from unnecessary explorations and thus improving the system’s robustness. Additionally, it minimizes the joint-wiggling process (see Section~\ref{sec:control}), which can be time consuming, because the robot only needs to evaluate motion along the known admissible motion direction, saving time and resources.

In order to rescue manipulation models, an active interconnection between control and planning components is required. This active interconnection takes the joint state from the planning component to adjust the manipulation trajectory and wiggling direction accordingly. This active interconnection improves the robustness by avoiding unnecessary explorations that affect the system’s efficiency and could result in manipulation failures.

\subsubsection{Active Grasp Pose Estimation}
\label{sec:graspPoseEstimation}

The perception component uses RGB images to detect joint handles and depth information to estimate their poses. An accurate estimation of handle poses is vital for the successful grasping of joints and their subsequent manipulation. However, two issues often hinder this process: perspective distortions and noisy depth measurements.

\begin{figure}[h]
	\centering
	\begin{subfigure}[b]{0.48\columnwidth}
		\centering
		\includegraphics[width=\textwidth]{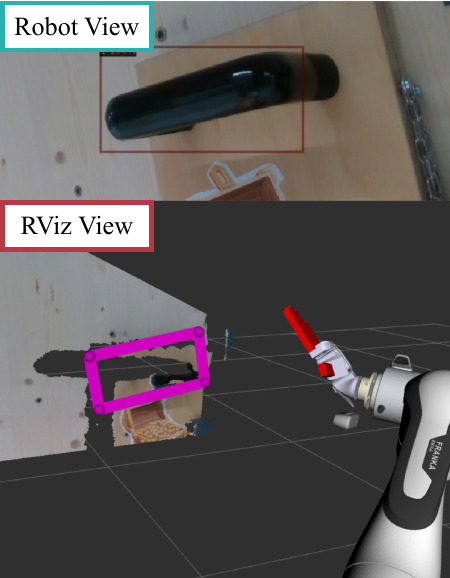}
		\caption{}
		\label{fig:bbox_resolution_orht_view}
	\end{subfigure}
	\hfill
	\begin{subfigure}[b]{0.48\columnwidth}
		\centering
		\includegraphics[width=\textwidth]{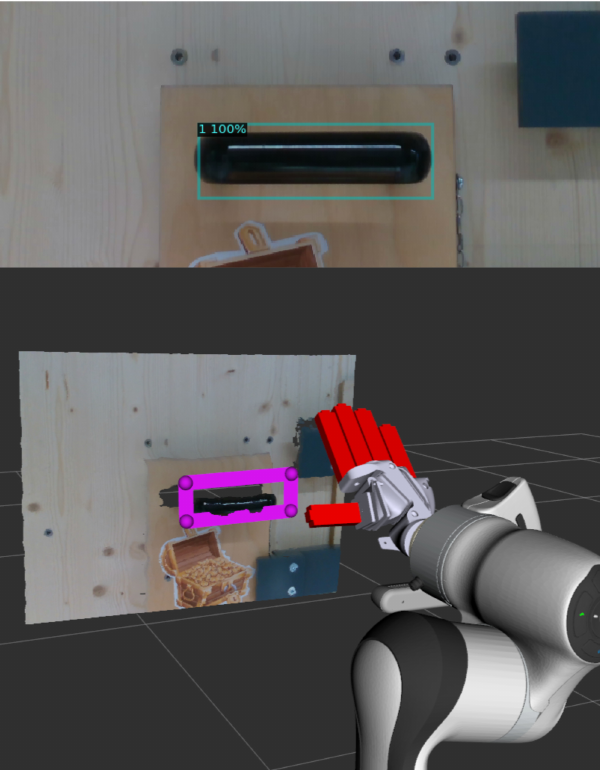}
		\caption{}
		\label{fig:bbox_resolution_tilted_view}
	\end{subfigure}
	\caption{Figure~$\mathbf{a}$ shows an imprecise handle pose estimate due to perspective distortion and noisy depth measurements. In contrast, the robot actively adjusts the joint handle's viewpoint by leveraging control in perception. This adjustment refines the handle pose estimation, leading to a more accurate result, as shown in Figure~$\mathbf{b}$.}
	\label{fig:bbox_resolution}
\end{figure}

We address these problems by establishing an active interconnection between perception and control components. This interconnection leverages the robot's ability to move its camera, revealing task-relevant visual information. The newly obtained visual data then guides further camera movements, creating a recursive cycle of information gathering that would be impossible to achieve without interconnecting components.

Specifically, this active interconnection exploits three key relationships between perception and control. First, after detecting handles, we center them in the image frame, which effectively minimizes perspective distortion. Second, as presented by~\cite{ling2024articulated}, moving the camera closer to a detected handle increases depth resolution, thereby reducing noise in the visual input. Third, inspired by~\cite{griffin2023grasppose}, we actively rotate the camera to identify a bounding box with minimal background interference, which is then used for more precise handle pose estimation.
By leveraging the active interconnection between control in perception, we obtain more accurate pose estimations, as visualized in Figure~\ref{fig:bbox_resolution}. 

Furthermore, this active interconnection allows the robot to effectively detect and track all handles within the lockbox, even with the limited field of view inherent to eye-in-hand setups. Overall, by actively interconnecting perception and control components, our system ensures more reliable detection of all handles and higher accuracy in the estimation of the corresponding grasp poses. This behavior contributes to the robustness of our system in solving lockboxes with varying configurations and handle positions and orientations.

\begin{figure*}[ht]
	\centering
	\includegraphics[width=\linewidth]{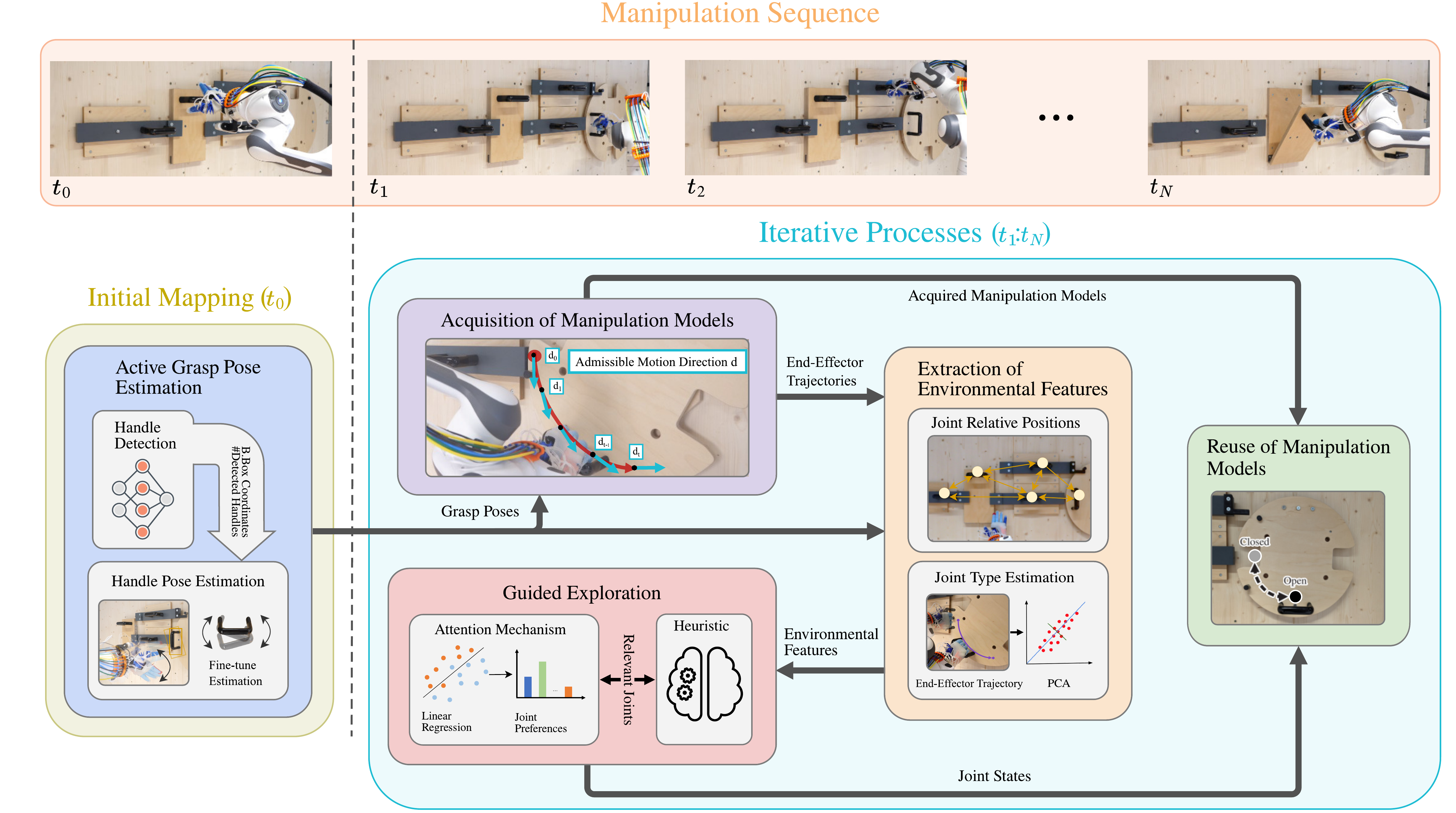}
	\caption{Behavior coordination for solving a lockbox. The components are depicted as circles in the bottom left, while the active interconnections between components are shown in the upper right corner of each behavior. Initially, the robot fine-tunes each detected handle pose using~\emph{Active Grasp Pose Estimation}. After that, the robot iteratively manipulates the lockbox’s joints. Specifically, when the robot encounters an unexplored joint, it wiggles the joint. If the joint is movable, it employs~\emph{Acquisition of Manipulation Models} to autonomously manipulate the joint. The resulting end-effector trajectory is stored in the planning component, allowing the robot to reuse this trajectory if the joint needs to be manipulated again. This end-effector trajectory is also used to estimate the joint type, which, together with the positions of joint handles, will be used by~\emph{Guided Exploration} to inform the planning component. The whole process of solving a lockbox can be seen in~\extensionone.}
	\label{fig:systemDiagram}
\end{figure*} 

\subsection{System Overview}
\label{sec:SystemOverview}

We now describe how the system coordinates all active interconnections to solve a lockbox. The solving process is illustrated in Figure~\ref{fig:systemDiagram} and~\extensionone. In the beginning, the robot follows a predefined trajectory to detect, track, and estimate the pose of each handle within a lockbox. It is necessary to perform such an active search behavior as we only use a wrist-mounted camera to receive vision information. Subsequently, the robot fine-tunes each handle pose using the~\emph{Active Grasp Pose Estimation} method and generates a grasp pose accordingly. These estimations, including the number of joints and handle poses, are used to initialize the planning and control components. After that, the robot iteratively manipulates the joints of the lockbox. When the robot encounters an unexplored joint, it first wiggles the joint. If the joint is movable, it autonomously manipulates the joint by following its articulation (\emph{Acquisition of Manipulation Models}). The resulting end-effector trajectory is stored in the planning component, allowing the robot to reuse this trajectory if the joint needs to be manipulated again (\emph{Reuse of Manipulation Models}). Note that this end-effector trajectory is also used to estimate the joint type, which, together with the positions of joint handles as environmental features, will be used by~\emph{Guided Exploration} to inform the planning component. 

\section{Experiments}
\label{sec:experiment}

To investigate the hypothesis that active interconnections between components enhance a robotic system's robustness, we adopt a systematic experimental approach. We start by introducing a minimally interconnected baseline system capable of solving the lockbox task. We then incrementally introduce additional active interconnections to create a series of systems. The baseline system (referred to as \textbf{Base}) incorporates essential active interconnections for solving lockboxes, including the \emph{Acquisition of Manipulation Models} with the wrench-gated trajectory generator  (Section~\ref{sec:ConstraintFollowing}) and the active exploration module for handle localization (Section~\ref{sec:perception}).

To distinguish between the various active interconnections, we employ a color-coded scheme for the associated components: \textbf{\textcolor{red}{P}} for \textbf{\textcolor{red}{perception}}, \textbf{\textcolor{blue}{C}} for \textbf{\textcolor{blue}{control}}, and \textbf{\textcolor{amber}{P}} for \textbf{\textcolor{amber}{planning}}. We now describe the robotic systems evaluated in our experiments,  characterized by varying degrees of active interconnections:

\begin{table*}[htp]
	\centering
	\small
	\begin{tabular}{@{}lcccccc@{}}
		\toprule
		& \multicolumn{3}{c}{Joint Interdependencies in Lockboxes} & \multirow{2}{*}{\makecell{No Prior\\Training Required}} & \multirow{2}{*}{\makecell{Autonomous\\Manipulation}} & \multirow{2}{*}{\makecell{Real-World\\Experiment}} \\
		\cmidrule(lr){2-4}
		& one-to-one & many-to-one & bistable-locking & & & \\
		\midrule
		Baum et al. \cite{baum2017opening} & \checkmark & \checkmark & --- & \checkmark & --- & \checkmark \\
		Verghese and Atkeson \cite{verghese2023using} & \checkmark & --- & --- & \checkmark & --- & \checkmark \\
		Ota et al. \cite{ota2023h} & \checkmark & \checkmark & --- & --- & \checkmark & --- \\
		Abbatematteo et al. \cite{abbatematteosensorized} & \checkmark & --- & --- & --- & \checkmark & --- \\
		Liu et al. \cite{liu2023busybot} & \checkmark & \checkmark & \checkmark & --- & \checkmark & \checkmark \\
		\textbf{Base} System (ours) & \checkmark & \checkmark & \checkmark & \checkmark & \checkmark & \checkmark \\
		\bottomrule
	\end{tabular}
	\vspace{1em}
	\caption{Our \textbf{Base} system, with a minimal amount of active interconnections, covers a greater spectrum of lockbox complexities than previous works. First, it employs a novel planning component that can handle lockboxes with different types of joint interdependencies. Second, our \textbf{Base} system can autonomously manipulate various mechanical joints by leveraging a perception-control active interconnection, as detailed in Section~\ref{sec:ConstraintFollowing}. Moreover, the \textbf{Base} system can solve various lockboxes without requiring prior training, enabling evaluation in diverse real-world settings.}
	\label{table:BaseSystemComparison}
\end{table*}

\begin{enumerate}	
	\item \baseSystem: the base system that contains only necessary component interconnections for solving lockboxes
	
	\item \weakPlanning: the base system with one additional interaction between perception and control (\textbf{\textcolor{red}{P}\textcolor{blue}{C}}) that allows the robot to perform~\emph{Active Grasp Pose Estimation}
	
	\item \weakControlPlanning: the base system with two active interconnections that allow the planning component to leverage the environmental features afforded by perception and control to perform ~\emph{Guided Exploration} (\textbf{\textcolor{red}{P}\textcolor{blue}{C}\textcolor{amber}{P}}) and the interconnection required for \emph{Active Grasp Pose Estimation} (\textbf{\textcolor{red}{P}\textcolor{blue}{C}})
	
	\item \weakPerceptionControl:~the base system with two additional interconnections for~\emph{Guided Exploration} (\textbf{\textcolor{red}{P}\textcolor{blue}{C}\textcolor{amber}{P}}) and~\emph{Reuse of Manipulation Models} (\textbf{\textcolor{blue}{C}\textcolor{amber}{P}})
	
	\item \richInteractions: system with the highest number of active interconnections among components as described in Section~\ref{sec:SystemOverview}
\end{enumerate}

It is worth noting that our \textbf{Base} system already covers a greater spectrum of lockbox complexities than previous works as summarized in~Table~\ref{table:BaseSystemComparison}. Concretely, ~\cite{baum2017opening, verghese2023using} mainly focused on the high-level reasoning of interlocking dependencies. Thus, they significantly implied the lockbox manipulation task, such as assuming manipulation policies for mechanical joints are known. ~\cite{ota2023h} presents a lockbox-solving framework that autonomously acquires manipulation policies for individual mechanical joints and reasons for the interlocking dependency. However, the proposed framework is only evaluated in simulation. Moreover, manipulation policies for mechanical joints are often trained using reinforcement learning algorithms, which require extensive interactions and environment resets, making it difficult to validate in real-world settings~\cite{ota2023h, liu23busybot, abbatematteosensorized}. By contrast, our \textbf{Base} system with a minimal degree of active interconnections has the ability to solve various lockboxes in the real world without the need for prior training. 

In the following sections, we will evaluate the performances of the aforementioned systems in both simulated and real-world environments. To assess system robustness, we introduced variations in lockbox scale, interlocking dependencies, robot pose, and joint type. Our experimental findings demonstrate a clear correlation between the amount of active interconnections and the system's robustness to these environmental changes.

\subsection{Active Interconnections Improve Planning Performance}
\label{sec:experiment:planning}

We first evaluate the impact of additional active interconnections on the lockbox-solving performance under variations in scale and interlocking dependencies of lockboxes in simulation. The simulated lockboxes share the same parameters (i.e. joint types and positions) with the physical lockbox but have two additional joints (\textbf{F} and \textbf{G}), as illustrated in Figure~\ref{fig:physicalLockbox}. Conducting this experiment in simulations allows us to test the robustness of the systems in solving large variations of lockboxes. For example, we can significantly increase the scale of the lockbox without considering the reachability-limitation that real robots have to deal with. We can also introduce more abstract lockbox-configurations that cannot be realized mechanically but can occur in electric puzzles.

We compare the lockbox-solving performance of three different systems. We assume these systems can physically manipulate each mechanical joint. The first system is \textbf{Base} which relies on our heuristic-based planning component to solve different lockboxes. The second system is \richInteractions~which assists the planning component with environmental features (i.e., joint types and positions) afforded by the active interconnection between perception and control. 

The third system is a Reinforcement Learning (RL) agent. This system employs a deep Q-Network (DQN) agent~\cite{mnih2015human} for planning. The DQN is trained using RL on randomized lockbox configurations for 10,000 episodes. The agent takes the joint states represented as a vector of binary values and the one-hot encoded index of the last attempted joint as input. The agent then predicts a quality value for each joint. The DQN architecture consists of a fully connected network with an input layer of size $2N$, with $N$ denoting the number of joints in the lockbox, a hidden layer with 64 units, and an output layer of size $N$. At each step, the agent receives a reward of $\Delta s \in \mathbb{Z}$, which is computed as the difference in the minimal number of remaining steps to solve the lockbox before and after manipulating a joint. Additionally, the agent receives a reward of 5 upon successfully solving the lockbox. Due to the fixed input size of the DQN, we have to train a distinct DQN agent for each lockbox scale (number of joints). Similar to the \textbf{Base} system, this baseline treats planning as a separate component, with no active interconnections to other components.

\begin{figure*}[htp!]
	\centering
	\includegraphics[width=\textwidth]{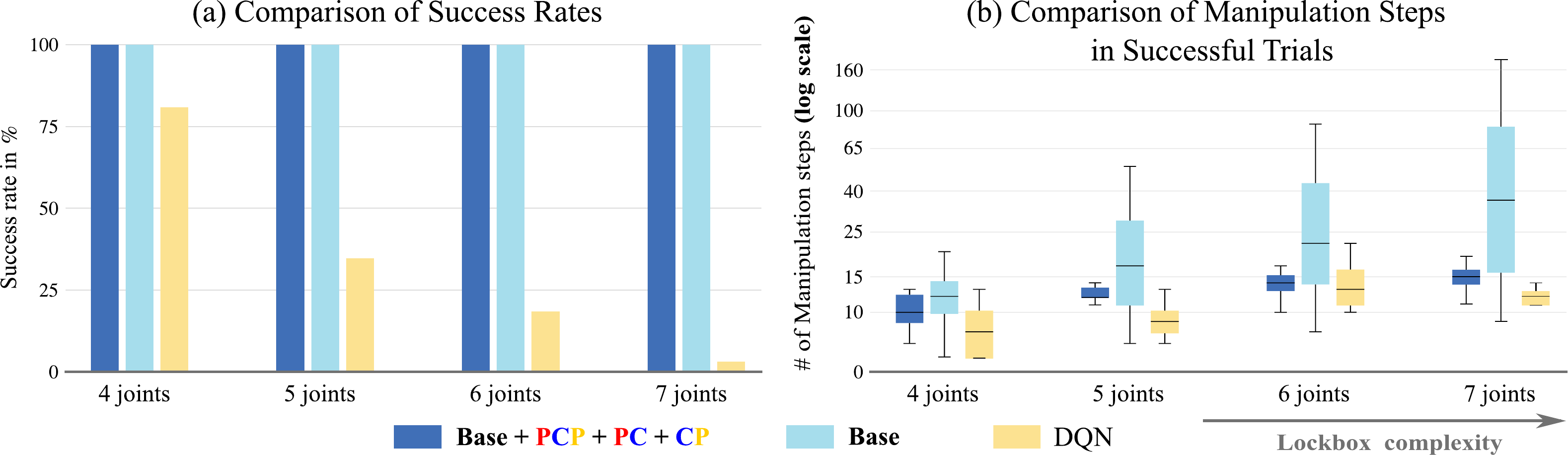}
	\caption{Performance comparison of three systems in solving simulated lockboxes of varying scales. In Figure~$\mathbf{a}$, the success rates of the three systems are compared at four lockbox scales. The distributions of the number of exhibited manipulation steps in successful trials for every system and lockbox scale are visualized in Figure~$\mathbf{b}$. While DQN required fewer manipulation steps in successful trials, it fails to generalize to unseen lockbox configurations, yielding a much lower success rate than the two other systems. By contrast, the~\textbf{Base} system reliably solved all lockboxes without the need for prior training. Moreover, our~\richInteractions~system with additional component interconnections demonstrates much more efficient lockbox-solving behaviors i.e., requires fewer manipulation steps. These results indicate that our~\textbf{Base} system is already capable of reliably solving lockboxes of varying scales at the symbolic level. More importantly, adding active interconnections enhances planning performance, reducing unnecessary manipulation steps and consequently improving the system's robustness.}
	\label{fig:results_sim}
\end{figure*}

We assess the performances across four different \textbf{scales} of lockboxes, spanning from 4 to 7 joints. The simulated lockboxes share the same joint positions, joint types, and interlocking dependencies with the physical lockbox. To maintain a generalizable evaluation of the DQN agent, which operates deterministically and would easily overfit to particular lockbox configurations, we randomize the joint labels (except for the target joint). This randomization is applied in each trial and for all systems. Each system undergoes evaluation through 1000 trials for each lockbox configuration. We define a trial as successful if the system succeeds at moving the target joint within 1000 manipulation steps. Figure~\ref{fig:results_sim} shows the success rates and required number of manipulation steps for these systems.

Figure~\ref{fig:results_sim}.\textbf{a} visualizes the impact of the three different systems on performance. We can clearly see that our \textbf{Base} and \richInteractions~systems achieve a 100\% success rate across all lockbox scales. In contrast, the DQN system requires extensive training for each lockbox scale and exhibits a significant decline in success rate as the number of joints increases. Notably, the DQN system only solves 3.2\% of the lockboxes with 7-joints.

This disparity in performance is primarily due to the DQN's limited generalization capability. The DQN can only solve a lockbox if it has previously encountered its specific interlocking dependencies during training. As the number of joints increases, the combinatorial explosion of potential dependencies renders it increasingly improbable for the DQN to have encountered all relevant configurations during training, leading to a decline in performance.
These findings support the observations in~\cite{verghese2023using} and~\cite{heo2023furniturebench} that RL struggles with extracting transferable policies, especially for long-horizon sequential manipulation tasks. Additionally, the substantial performance gap between the DQN and the other systems underscores the complexity of our lockbox problem, particularly due to the presence of bistable-locking joints as discussed in Section~\ref{sec:lockboxenvironment}.

In addition to success rates, Figure~\ref{fig:results_sim}.\textbf{b} also highlights that the \richInteractions~system requires significantly fewer manipulation steps compared to the \textbf{Base} system. This gap widens as the lockbox complexity increases. The substantial difference in manipulation steps indicates that the planning component benefits greatly from the guidance provided by the active interconnection between the perception, control, and planning components. This active interconnection facilitates more efficient exploration by reducing unnecessary attempts that could lead to failures and consequently contributing to more robust behaviors. This result confirms the contribution of active interconnections to the robustness against various lockbox scales. 

\begin{figure}[htp]
	\begin{subfigure}[b]{\columnwidth}
		\centering
		\includegraphics[width=\columnwidth]{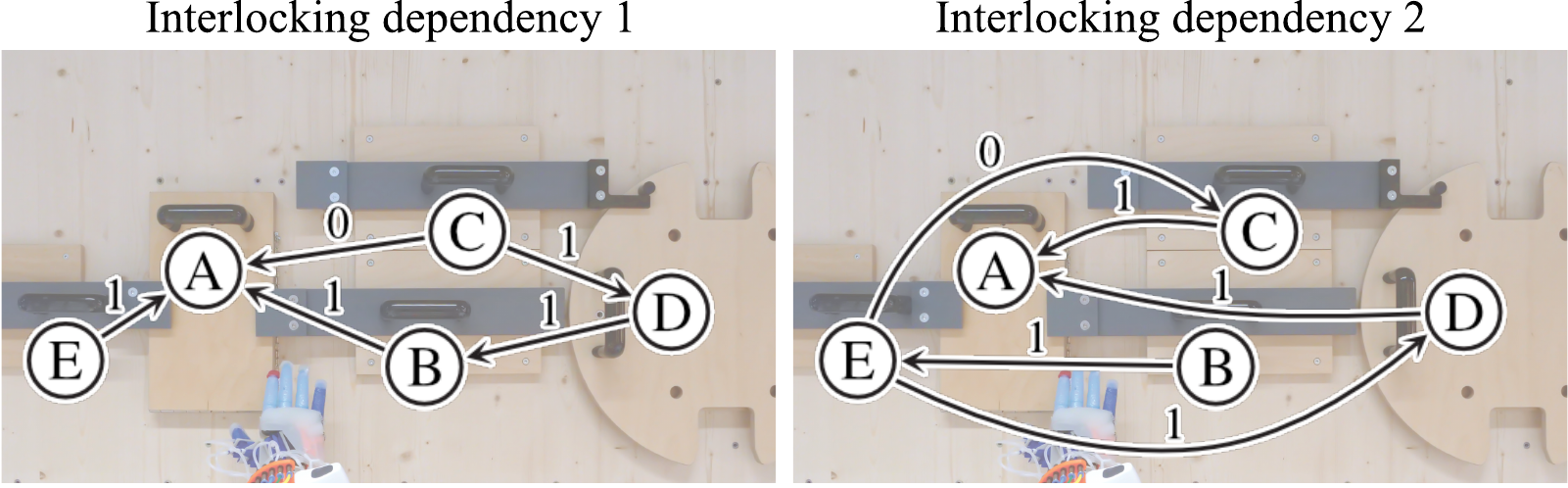}
		\caption{Two distinct lockbox configurations. In Interlocking Dependency 1 (left), a joint is more likely to be locked by nearby joints. Conversely, in Interlocking Dependency 2 (right), a joint is more likely to be locked by distant joints. These distinct configurations are used to test the \richInteractions~system's ability to adapt to different interlocking patterns.}
		\label{fig:newConfiguration}
	\end{subfigure}
	\begin{subfigure}[b]{\columnwidth}
		\centering
		\includegraphics[width=\columnwidth]{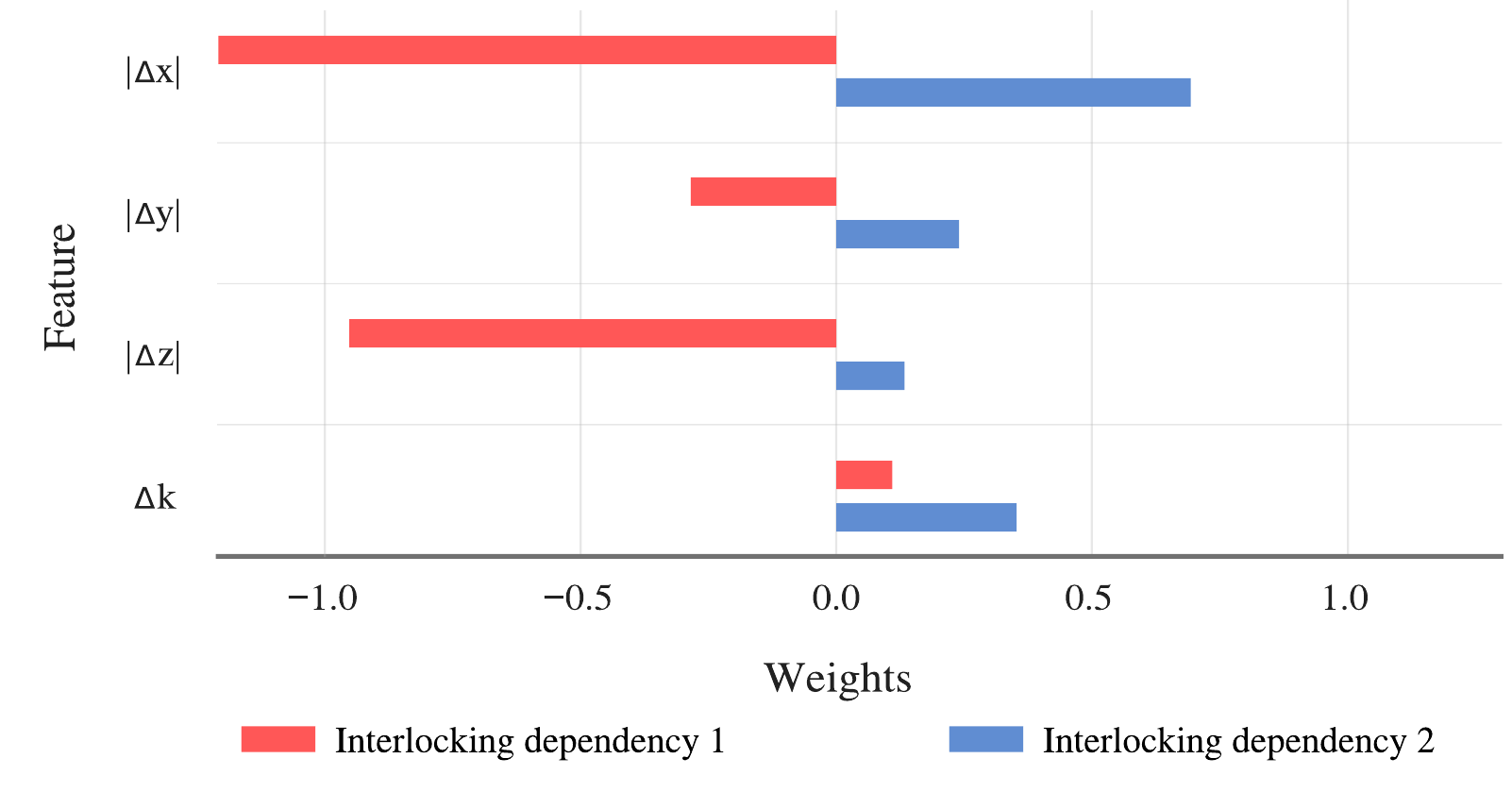}
		\caption{Visualization of the adapted weights of the attention mechanism for both configurations. Blue bars represent Interlocking Dependency 2 (distant locking), while red bars represent Interlocking Dependency 1 (nearby locking). Notably, the difference in weight signs for distance features $\left(|\Delta x|, |\Delta y|, |\Delta z|\right)$ highlights the mechanism's ability to adapt to different configurations with distinct interlocking patterns.}
		\label{fig:onlineWeights}
	\end{subfigure}
	\caption{Our attention mechanism captures different interlocking patterns, contributing to the system's robustness to environmental variations.}
\end{figure}

We now examine whether active interconnections can also contribute to robustness against variations in the  \textbf{interlocking dependency}. To do so, we designed a fictive configuration (Interlocking Dependency 2), as shown in Figure~\ref{fig:newConfiguration}. 
In this new configuration, joints are more likely to be locked by distant joints, rather than nearby joints, as in the previous configuration (Interlocking Dependency 1). We test whether the \richInteractions~system with the attention mechanism is still able to improve the planning efficiency in this new configuration. We ran 1,000 trials for both \textbf{Base} and \richInteractions~systems and compared the required number of manipulation steps to solve the lockbox and the learned weights of the attention mechanism. 

The \richInteractions~system solves the lockbox with Interlocking Dependency 2 within an average of 8.9 manipulation steps, whereas the \textbf{Base} system averages at 12.64 steps. It shows that the attention mechanism afforded by the active interconnection between perception, control, and planning improves the planning efficiency by 29.6\%. This improvement can be explained by the learned weights, as visualized in Figure~\ref{fig:onlineWeights}. In Interlocking Dependency 1 (matching the physical lockbox), the attention mechanism assigns negative weights to the features $\left(|\Delta x|, |\Delta y|, |\Delta z|\right)$, prioritizing joints \emph{closer} to the last attempted joint. In Interlocking Dependency 2, the weights become positive, indicating that joints \emph{farther} from the last attempted joint are prioritized. The difference in learned weights shows the attention mechanism's ability to extract appropriate inductive biases through online weight adaptation, thus enhancing planning performance across various lockbox configurations.

Overall, our experiments demonstrate that the \richInteractions~system characterized by a high degree of active interconnections exhibits high adaptability and efficiency in solving lockboxes across diverse environmental conditions, including variations in scale and configuration. Notably, this enhanced robustness is attained without significant modifications to individual system components but through the strategic establishment of active interconnections between them.

\subsection{Active Interconnections Enable Robust Real-World Manipulation}
\label{sec:experiment:realworldExperiments}

Transitioning to real-world manipulation, our experiments further substantiate the positive correlation between active interconnections and system robustness. In the first part, we observe a clear correlation between the level of active interconnections and robustness in real-world manipulations. Subsequently, we demonstrated that the \richInteractions~system can effectively generate robust manipulation behaviors for solving lockboxes with diverse joint types and poses, and lockbox scales.

\begin{figure}[htp]
	\centering
	\includegraphics[width=\columnwidth]{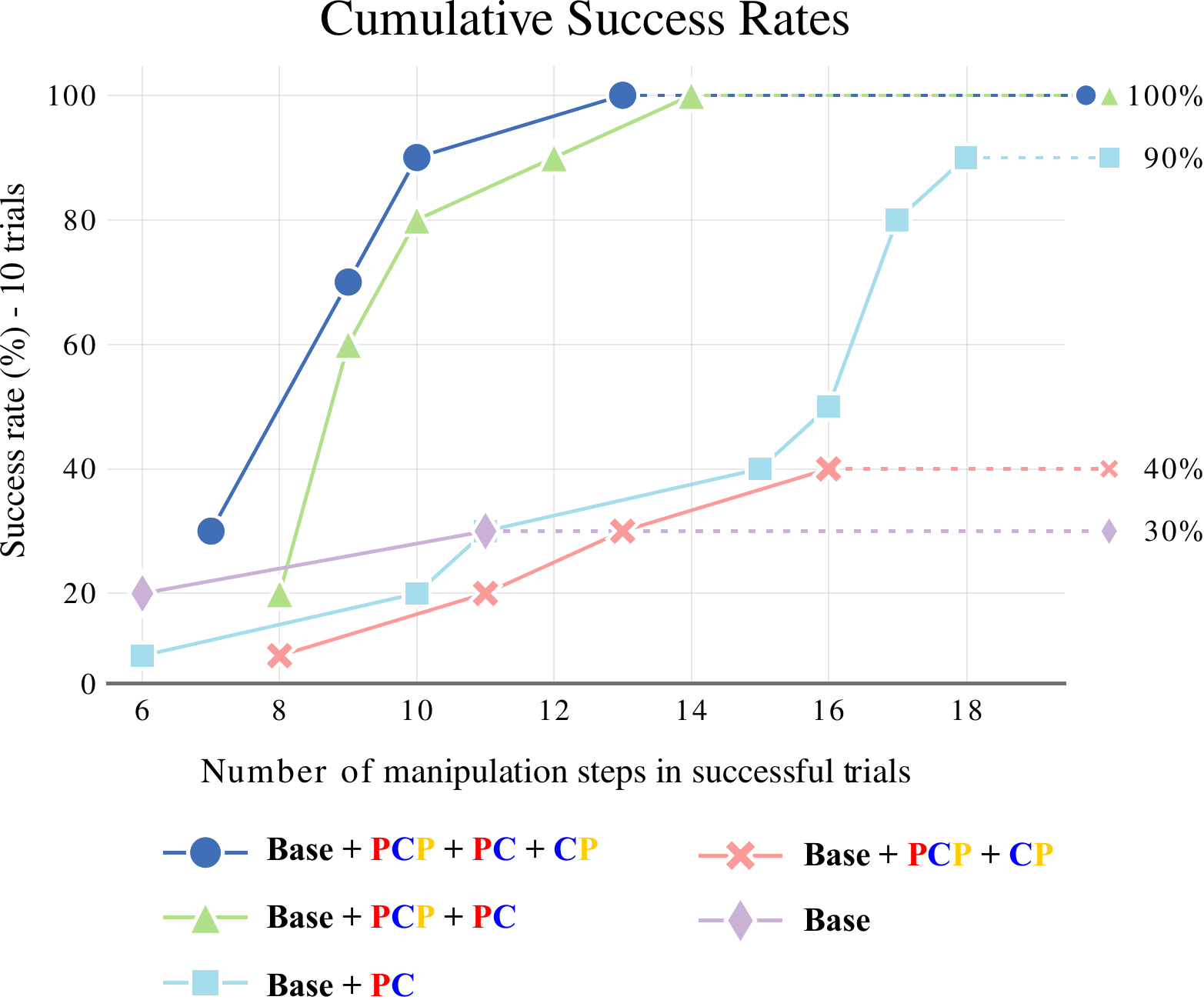}
	\caption{Cumulative success rates of solving the lockbox as a function of manipulation steps for different system configurations. Each point on the figure represents the probability that the corresponding system successfully solves the lockbox within the number of steps indicated by its x-coordinate. For example, \richInteractions~could solve 9 of the 10 trials requiring 10 manipulation steps or less. In contrast \weakPlanning~could only solve 2 of the 10 trails for that same steps limit. In this plot, higher lines reflect a higher success rate, while the line shifted towards the left reflects a higher efficiency i.e., requiring fewer manipulation steps. We can see that increasing amounts of active interconnections (from \textbf{Base} to \richInteractions) leads to \textit{higher} success rates in solving the lockbox while requiring \textit{fewer} manipulation steps.}
	\label{fig:real_world_sr}
\end{figure}

To evaluate how our biology-inspired design principle affect the system’s robustness, we gradually add active interconnections to the \textbf{Base} system and test the performance of the subsequent systems on the physical lockbox with 4 joints (joint $\mathbf{A}$, $\mathbf{B}$, $\mathbf{C}$, $\mathbf{D}$). We evaluated each of the five systems, described in Section~\ref{sec:experiment}, for 10 trials. Each trial was considered successful if the robot managed to open the target joint within 20 manipulation steps. We evaluated the success rate (percentage of successful trials) and the number of interactions required by the robot to solve the lockbox. The results are presented in Figure~\ref{fig:real_world_sr} as cumulative success rates as a function of the number of manipulation steps.

\subsubsection{Success Rate}
\label{sec:experiment:realworldExperiments:succeesRate}

We first analyze the success rates, as shown in Figure~\ref{fig:real_world_sr}. The results show a clear distinction in performance between the systems. The \textbf{Base} system succeeded 3 out of 10 times in solving the lockbox. By adding the active interconnection \textbf{\textcolor{red}{P}\textcolor{blue}{C}} (perception and control) to the \weakPlanning~system, we observe a significant increase in the success rate (90\%). Similarly, adding the active interconnections \textbf{\textcolor{red}{P}\textcolor{blue}{C}\textcolor{amber}{P}} and \textbf{\textcolor{blue}{C}\textcolor{amber}{P}} further increased the success rates to 100\%. 

\begin{figure}[htp]
	\centering
	\includegraphics[width=\columnwidth]{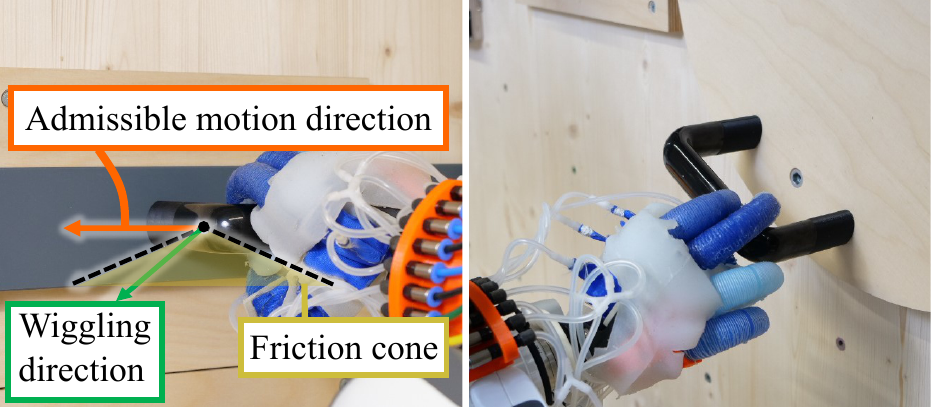}
	\caption{Two major failures caused by inaccurate grasp poses. First, inaccurate grasp poses can hinder the robot's "wiggling" motion within the friction cone (left). This can lead to misinterpretations of joint state and consequently failed manipulation. Second, a poor grasp can cause the robot to lose contact with the handle, particularly for joint \textbf{D}, which requires a large motion range (right).}
	\label{fig:failureModes}
\end{figure}

This difference in success rates is mainly attributed to the active interconnection \textbf{\textcolor{red}{P}\textcolor{blue}{C}} (between perception and control), which enables the robot to perform \emph{Active Grasp Pose Estimation}, improving the grasp pose estimation (see Section~\ref{sec:graspPoseEstimation}. Having accurate estimations of grasp poses for good grasps is crucial for robust manipulation. This is because the robot has to perform extensive forceful interactions with the mechanical joints to acquire their locking-state (locked or unlocked) and their manipulation policies. A reliable grasp is required for a successful policy acquisition. Without the active interconnection \textbf{\textcolor{red}{P}\textcolor{blue}{C}}, two failure modes can occur, as illustrated in Figure~\ref{fig:failureModes}. 

Similar to the active interconnection \textbf{\textcolor{red}{P}\textcolor{blue}{C}}, the success rates is further improved by \textbf{\textcolor{red}{P}\textcolor{blue}{C}\textcolor{amber}{P}}, which improves planning by reducing unnecessary explorations that could otherwise lead to failures. Overall, this experiment highlights the critical role of active interconnections, especially between perception and control, for achieving robust real-world manipulation behaviors. As can be seen in this~\extensionone, the \richInteractions~system with the highest amount of active interconnections was able to repeatedly solve the lockbox 10 out of  10 times over a three-hour period, even with an impaired finger of the end-effector midway through the experiment. These results clearly indicate the positive correlation between levels of active interconnections with the robustness of real-world manipulation behaviors.  

\subsubsection{Planning Performance}
\label{sec:experiment:realworldExperiments:efficiency}

As mentioned in the previous section, active interconnections also improves planning when solving lockboxes. It is showcased in Figure~\ref{fig:real_world_sr} that the \richInteractions~and \weakControlPlanning~systems require significantly fewer manipulation steps than others thanks to the active interconnection \textbf{\textcolor{red}{P}\textcolor{blue}{C}\textcolor{amber}{P}}. This aligns with our previous simulated experiments, suggesting that the planning component benefits from leveraging environmental features provided by the active interconnection between perception, control, and planning, leading to more efficient lockbox-solving behaviors.

Interestingly, \textbf{\textcolor{blue}{C}\textcolor{amber}{P}} does not have a substantial impact on the system performance, because both \richInteractions~and \weakControlPlanning~systems achieve similar performance. We attribute this observation to the use of a soft end-effector. Its inherent compliance and large contact area simplify the control problem, enhancing overall robustness and potentially diminishing the impact of the control-planning interconnection. This finding also indicates that not all active interconnections contribute to a system's robustness equally.

\subsubsection{Robustness to Environmental Variations}
\label{sec:experiment:realworldExperiments:variations}

The previous experiment clearly indicates that a system with more active interconnections exhibits more robust real-world manipulation behaviors that manifest themselves in high success rates and improved planning efficiency. In this experiment, we test the \richInteractions~system, which has the highest amount of active interconnections, in response to variations in lockbox scale, poses, as well as the morphology of the end-effector.

\begin{description}[style=unboxed,leftmargin=0cm]
	\item[Robustness to Variations in Lockbox-Scale:] The robustness of our \richInteractions~system extends to a larger lockbox with 5 joints without any modifications to the system (see~\extensionfour~and~\extensionfive). The system achieved a 100\% success rate (10 out of 10 trials), with an average of 13.5 manipulation steps per trial. It is important to note that the DQN agent (in simulation) managed to solve only 30\% of the trials on this scale of the lockbox. As we explained earlier, this robustness is attributed to the different active interconnections between the system's components, which enable efficient planning and robust real-world manipulation behaviors. Note that we could not significantly vary the pose of the robot in the experiment involving the 5-joint lockboxes as we could do with the 4-joint lockboxes. This is due to the limited manipulation range allowed by our current robot platform. To address this limitation, we are working on integrating the robot platform into a mobile base.
	
	\item[Robustness to Pose Changes:] First, we validate the system's robustness to pose changes (i.e., generalizes to new lockbox poses). We conducted 10 trials with different robot base poses within the robot's joint limits, as shown in Figure~\ref{fig:pose_variations_exp} and~\extensiontwo. Our \richInteractions~system successfully maintained its performance, solving the lockbox in all 10 trials. This robustness stems from the active interconnection between perception and control, allowing the robot to actively search for and accurately locate handles of mechanical joints in the lockbox.
	
	\begin{figure}[htp]
		\centering
		\includegraphics[width=\columnwidth]{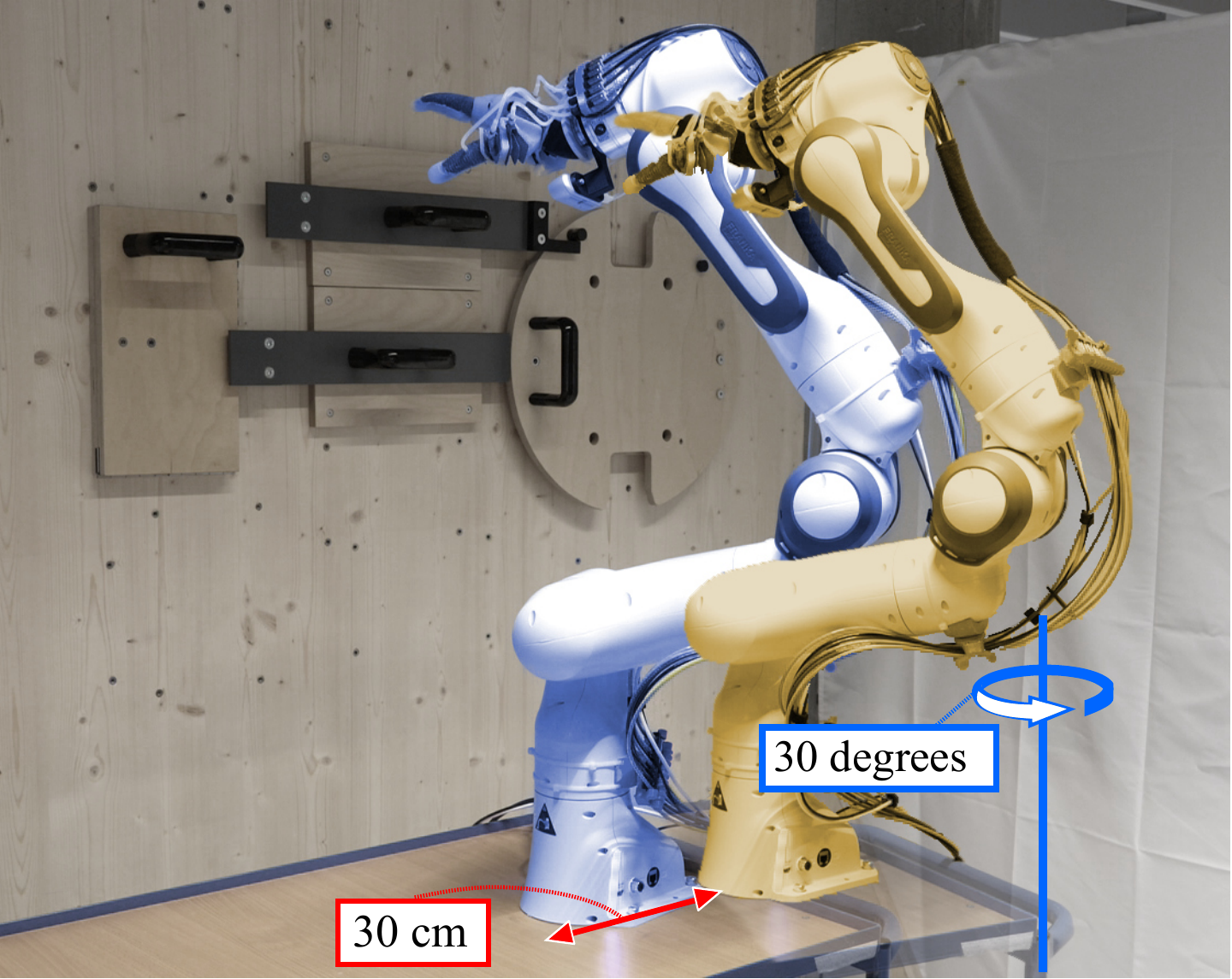}
		\caption{Our robot operating at different poses. We change the robot's initial pose by up to 30 cm and 30 degrees to introduce variations in the poses of mechanical joints, illustrated by the discrepancy between the blue and yellow robot poses. Our \richInteractions~system successfully maintained task performance despite these variations.}
		\label{fig:pose_variations_exp}
		\vspace{-1em}
	\end{figure}
	
	\item[Robustness to End-Effector Impairment:] Next, we demonstrate the ability of our \richInteractions~system to adapt to changes in the morphology of the end-effector caused by finger impairment. We did this by disabling certain fingers of the end-effector and running the system to solve the lockbox. Our \richInteractions~system accomplished the task with 6 morphological variations, as illustrated in Figure~\ref{fig:fingerImpairments} and~\extensionthree. This robustness stems from using the wrench-gated trajectory generator, enabled by the active interconnection between perception and control (see Section~\ref{sec:ConstraintFollowing}). This active interconnection allows the system to compliantly manipulate the joints within a desired wrench limit, as visualized in Figure~\ref{fig:constraintFollowingForce}. By setting a wrench limit that suffices with three fingers, the system can exploit the redundancy of the end-effector and remain robust with up to two fingers being impaired.
	
	
	\newlength{\subfigwidth}
	\setlength{\subfigwidth}{0.32\columnwidth}
	
	\newlength{\cropleft}
	\newlength{\cropright}
	\newlength{\croptop}
	\newlength{\cropbottom}
	\setlength{\cropleft}{220pt}
	\setlength{\cropright}{160pt}
	\setlength{\croptop}{50pt}
	\setlength{\cropbottom}{70pt}

	\begin{figure}[htp]
		\centering
		\begin{subfigure}[b]{\subfigwidth}
			\centering
			\includegraphics[width=\textwidth,trim={\cropleft} {\cropbottom} {\cropright} {\croptop},clip]{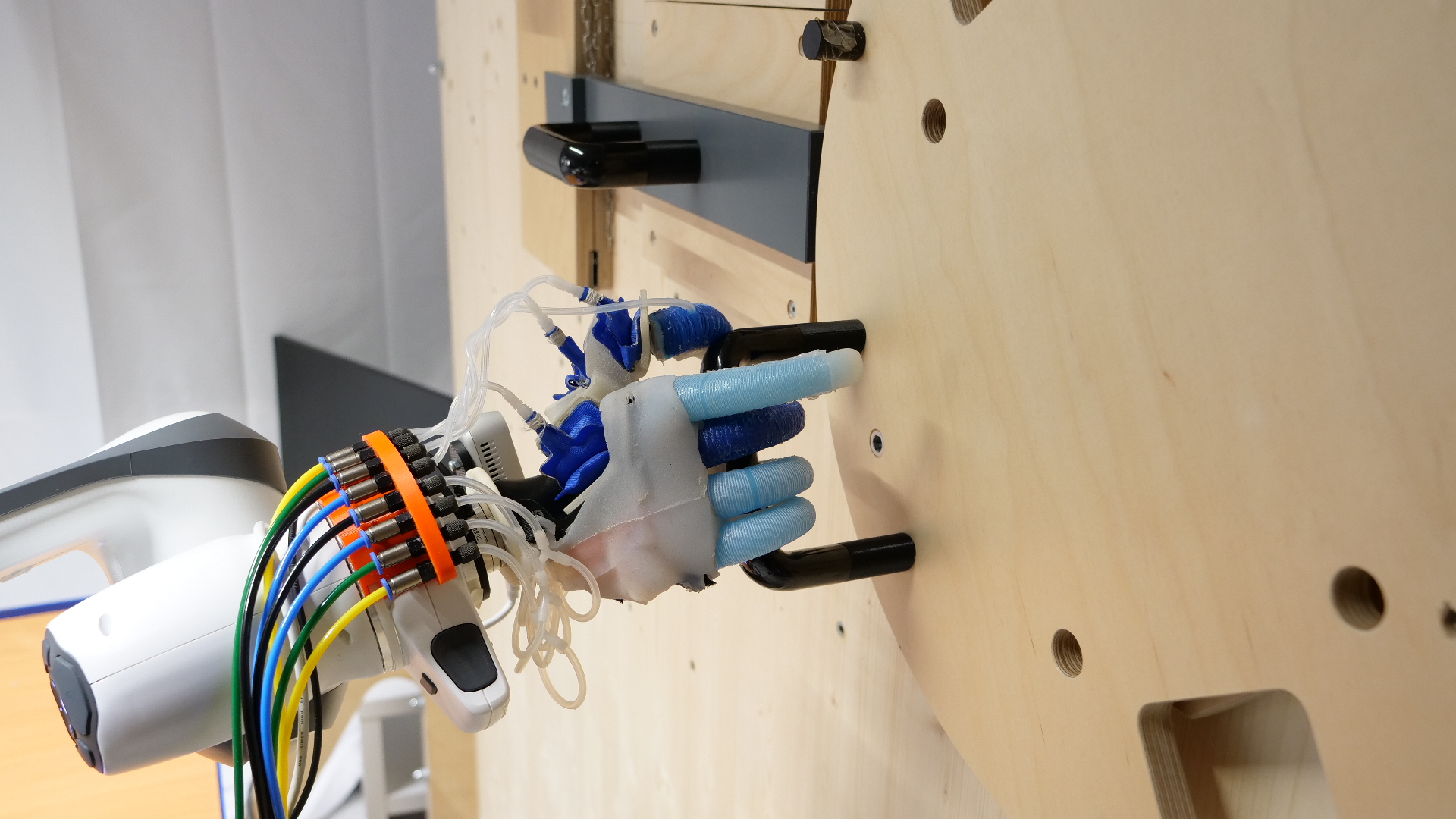}
			\caption{Index finger \\\hspace{\textwidth} deactivated}
		\end{subfigure}
		\hfill
		\begin{subfigure}[b]{\subfigwidth}
			\centering
			\includegraphics[width=\textwidth,trim={\cropleft} {\cropbottom} {\cropright} {\croptop},clip]{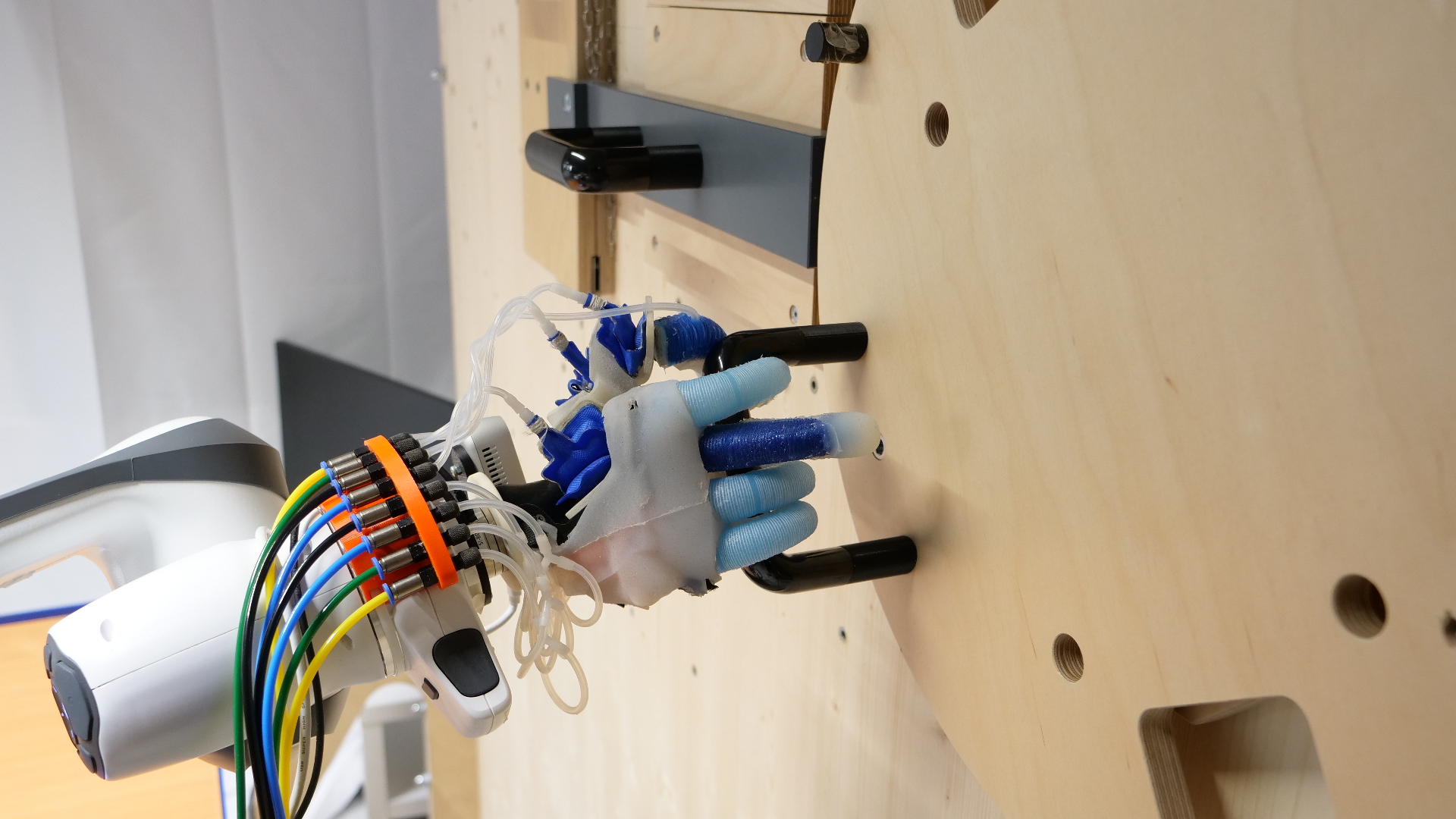}
			\caption{Middle finger \\\hspace{\textwidth} deactivated}
		\end{subfigure}
		\hfill
		\begin{subfigure}[b]{\subfigwidth}
			\centering
			\includegraphics[width=\textwidth,trim={\cropleft} {\cropbottom} {\cropright} {\croptop},clip]{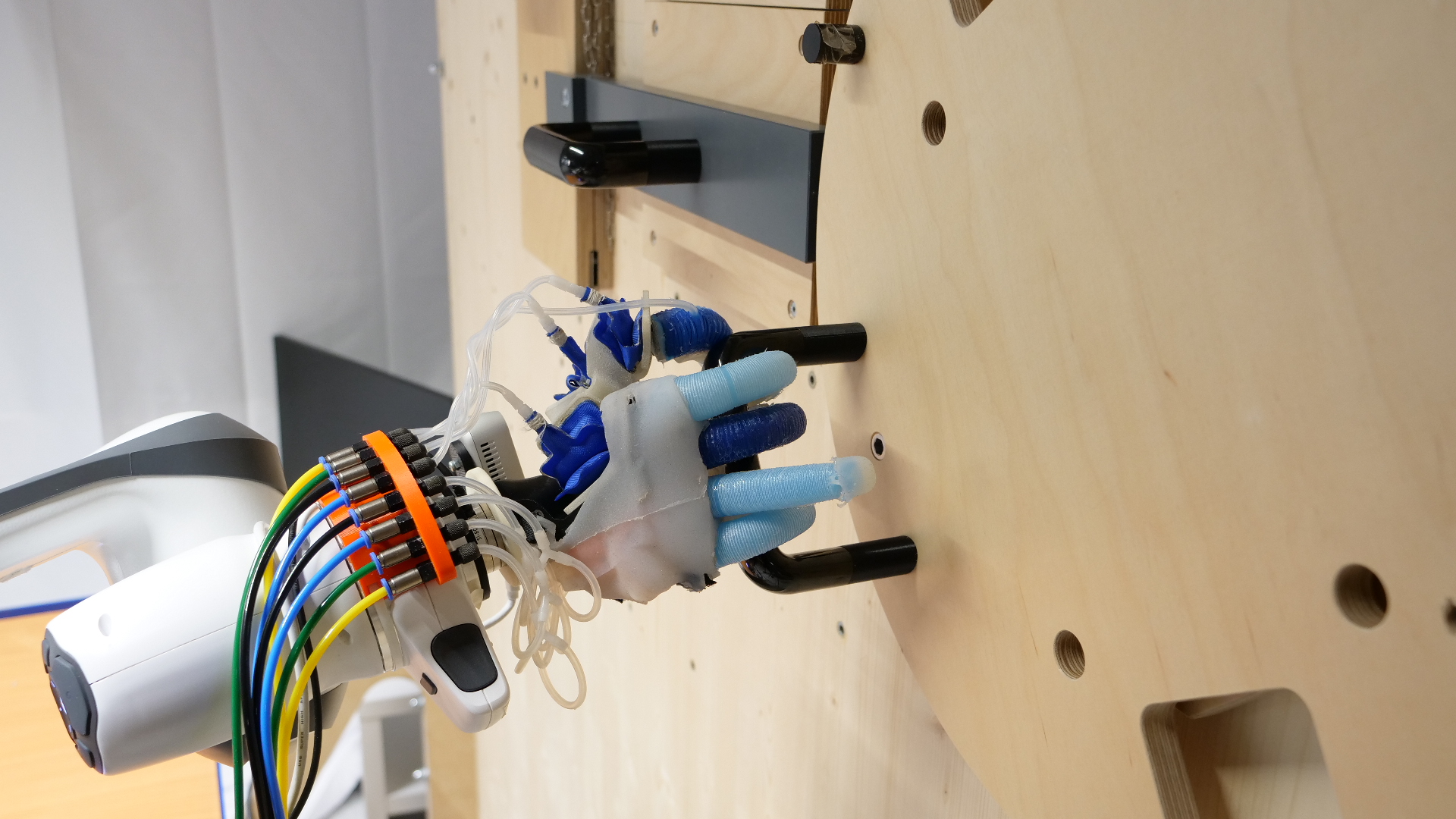}
			\caption{Ring finger \\\hspace{\textwidth} deactivated}
		\end{subfigure}
		
		\begin{subfigure}[b]{\subfigwidth}
			\centering
			\includegraphics[width=\textwidth,trim={\cropleft} {\cropbottom} {\cropright} {\croptop},clip]{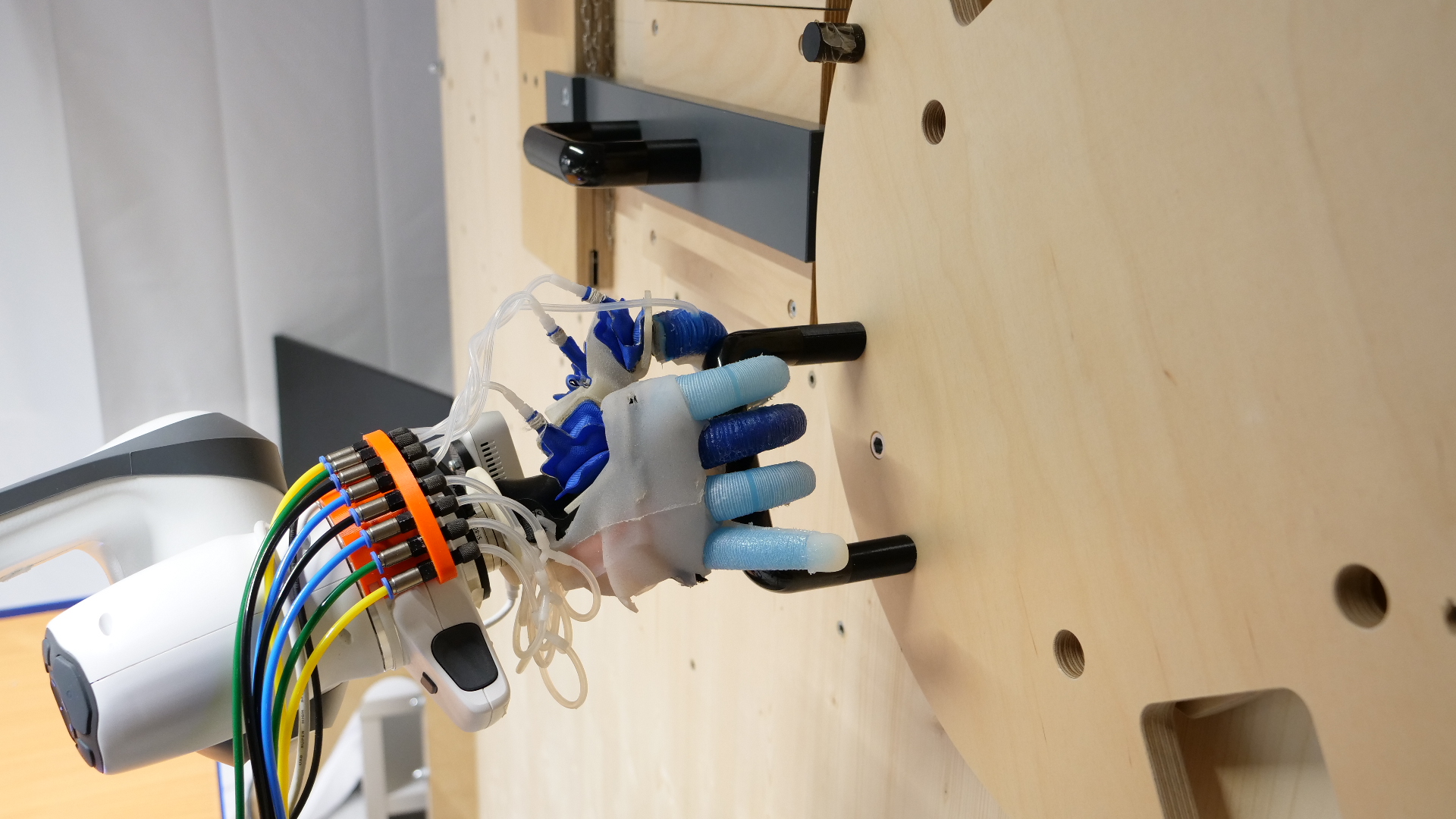}
			\caption{Little finger \\\hspace{\textwidth} deactivated}
		\end{subfigure}
		\hfill
		\begin{subfigure}[b]{\subfigwidth}
			\centering
			\includegraphics[width=\textwidth,trim={\cropleft} {\cropbottom} {\cropright} {\croptop},clip]{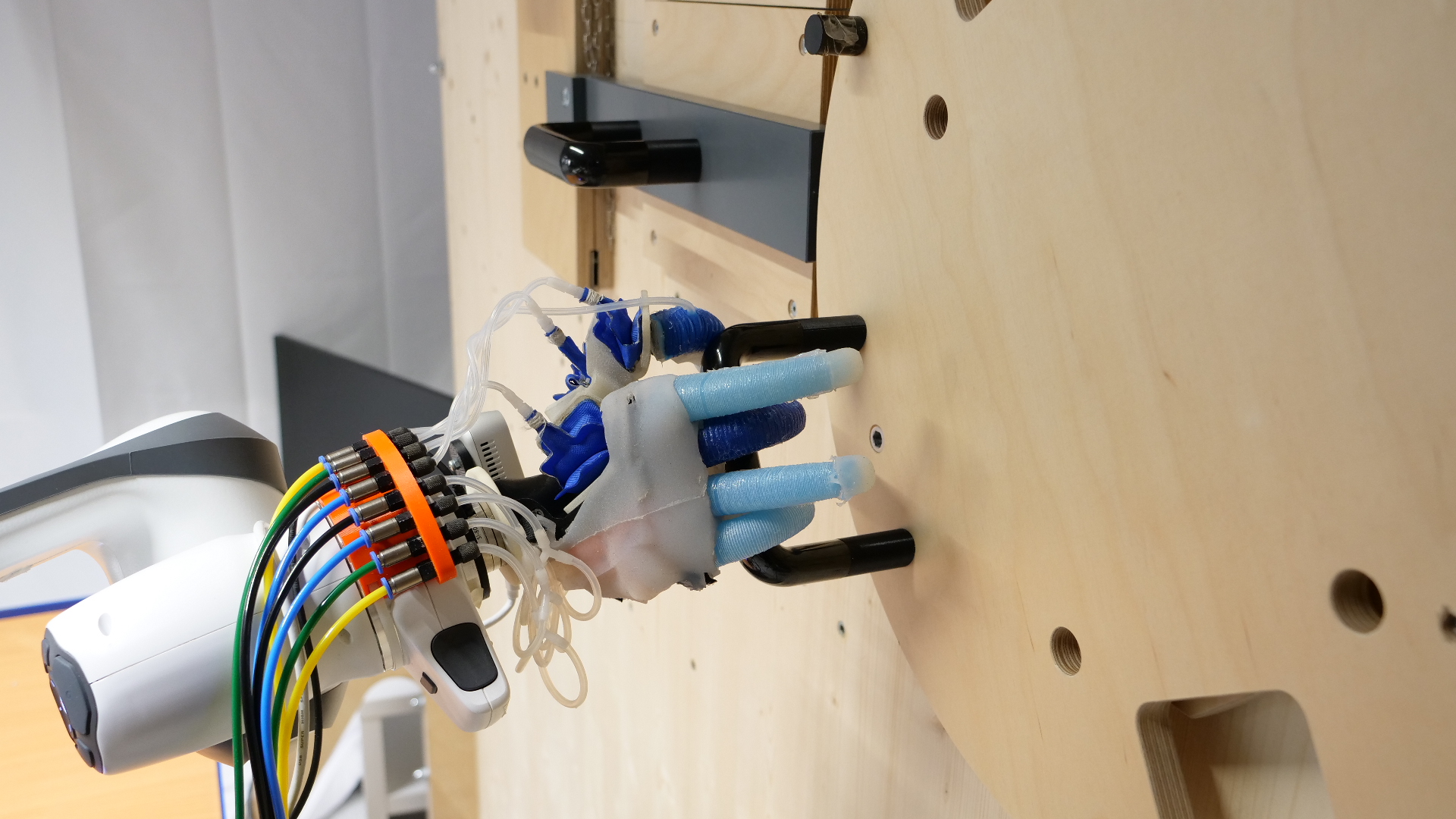}
			\caption{Index and ring \\\hspace{\textwidth} fingers deactivated}
		\end{subfigure}
		\hfill
		\begin{subfigure}[b]{\subfigwidth}
			\centering
			\includegraphics[width=\textwidth,trim={\cropleft} {\cropbottom} {\cropright} {\croptop},clip]{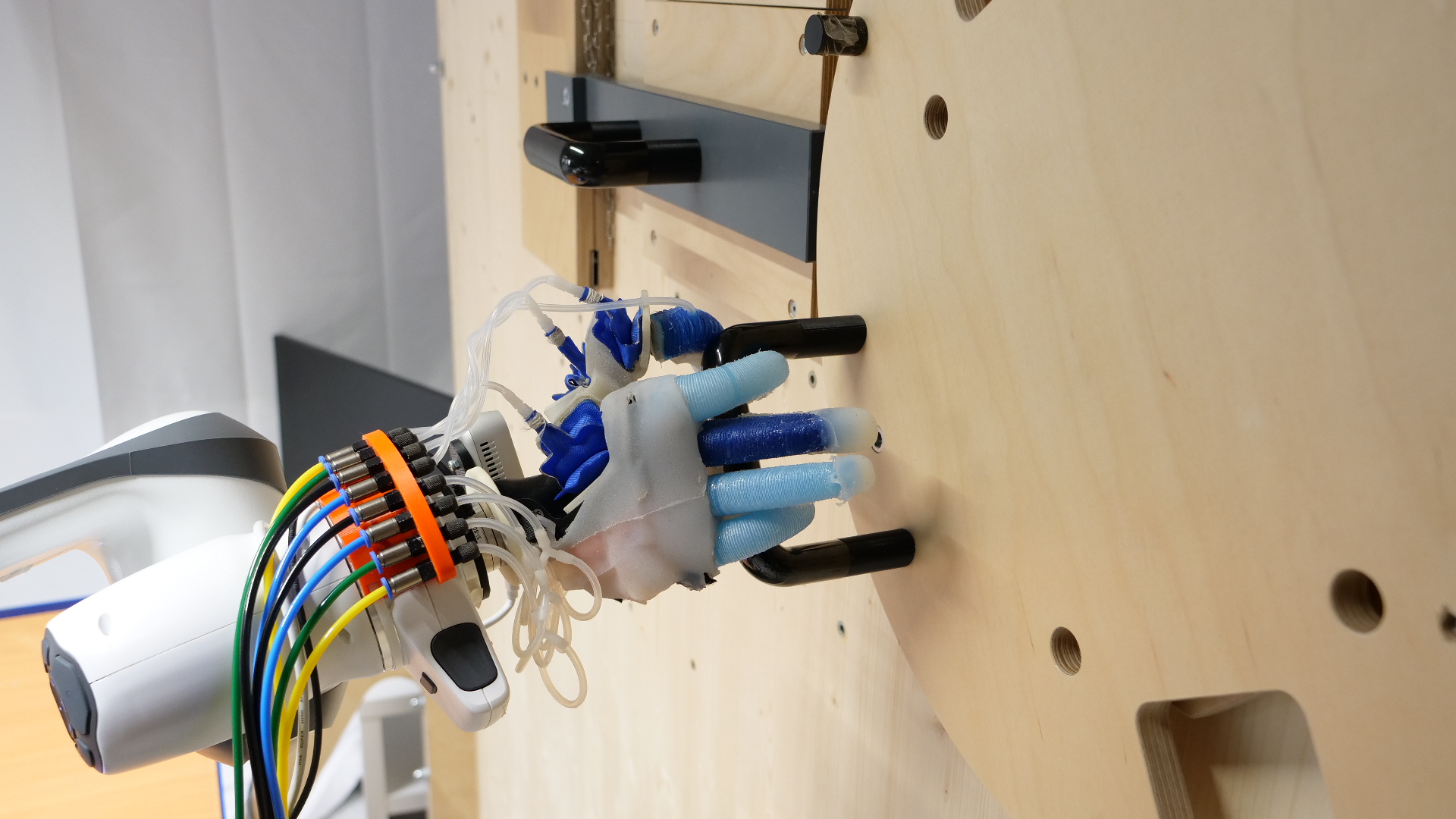}
			\caption{Index and ring \\\hspace{\textwidth} fingers deactivated}
		\end{subfigure}
		\caption{The \richInteractions~system is robust to different end-effector morphologies caused by finger impairments.}
		\label{fig:fingerImpairments}
	\end{figure}
	
\end{description}

Our findings underscore the system's capacity to adapt to diverse environmental conditions without fundamental structural modifications. Crucially, these results emphasize that fostering active interconnections among system components is crucial for achieving robust robotic behavior. This approach diverges from the traditional focus on individual component sophistication. By exploring this new design paradigm, we identify a promising avenue for enhancing behavioral diversity, a critical factor in adaptability in complex environments.

\subsection{Limitations in the Lockbox Solving System}
\label{sec:limitations}

The real-world experiments have demonstrated that a system with sufficient active interconnections can reliably solve lockboxes with various scales, interlocking dependencies, poses, and types of mechanical joints, showcasing substantial robust manipulation behaviors. However, our experiments have also highlighted some limitations, indicating a need for more active interconnections among the system's components.

\begin{description}[style=unboxed,leftmargin=0cm]
	\item[Limited Affordance Prediction:] Our physical lockbox uses standard door handles, simplifying the affordance prediction problem. However, real-world scenarios involve diverse elements. A desired extension is to incorporate an affordance prediction model~\cite{chen2023survey}. As suggested in~\cite{hausman2015active, zhao2024tac, ling2024articulated}, the robot must refine the affordance prediction results through active exploration, which highlights the need for additional active interconnections between perception and control.
	
	\item[Handling Multi-DoF Joints:] Our current system assumes all joints being single-DoF, simplifying the manipulation problem. For multi-DoF joints, a new system component is necessary to learn suitable manipulation policies. \cite{tahara2022disturbance} and \cite{li2023augmentation} propose such components that learn from human demonstrations. Importantly, they further refine initial policies by incorporating exploration actions to discover states not covered in the demonstrations, showing advantageous interconnections between learning and other system components, such as perception and control.
	
	\item[Accounting for Uncertainty and Non-Determinism:] The system currently assumes binary states for all joints and a deterministic environment. However, the interaction with a lockbox in the real world is probabilistic, failures could occur if joint states deviate from the binary state assumptions due to errors in perception and actuation. This highlights the need for a failure recovery mechanism \cite{paxton2019representing, baum2022world}, necessitating more intricate active interconnections between the perception, control, and planning components to effectively detect erroneous positions of mechanical joints and autonomously recover from such failure situations.
	
\end{description}

In summary, these limitations suggest even more active interconnections between system components to further enhance the robustness against a boarder range of environmental variability in the lockbox manipulation task. 

\section{Discussion}
\label{sec:discussion}

Our experimental results demonstrate that the proposed design principle of active interconnections enables robustness in robotic systems. The example system for opening lockboxes continued to perform successfully under significant environmental changes, including variations of the lockbox in scale, interlocking dependencies, joint types, and relative lockbox/robot poses. Performance is also robust to defects in the robot's end-effector. These results support our hypothesis that active interconnections contribute to robustness in robotic systems. Based on these findings, we will now speculate on why this is the case.

\subsection{Regularities}

Closed-loop feedback controllers are robotics' workhorse when it comes to achieving robustness in real-world systems. The key to robust control is the choice of an appropriate control function, describing the mapping from input to output of the controller. Control functions lead to robust system behavior when they represent a \textit{regularity} of the system they control.

A \textit{regularity} is a reproducible and predictable relationship between the robot's embodiment, features of the environment, and the behavioral consequences of their interactions.  A regularity limits the degrees of freedom in the space in which the control function is described. This function can then be viewed as a lower-dimensional manifold in that space. Regularities capture a property of the world that is relevant to the robot. They facilitate the rejection of disturbances, aid with the interpretation of percepts, and simplify the selection of actions. In short, regularities are the foundation of robustness in robotic systems.

The Jacobian matrix of a robotic manipulator is an example of regularity. It captures the relationship between changes in joint angles and changes in end-effector pose.
If a task requires the control of the end-effector via the joint variables, exploiting this regularity improves robustness. There are many examples of robots exploiting regularities, such as performing guarded moves, achieving force closure, or servoing to a visual target.

Other disciplines also rely on regularities. In machine learning, for example, regularities are sometimes called \textit{inductive biases} or \textit{priors}. Representation learning can be understood as the automated attempt to extract regularities from data.

\subsection{Active Interconnections Compose Regularities}

When designing robotic systems, roboticists endow their systems with the ability to exploit regularities. These regularities occur in all components of the system: planning, control, perception, reasoning, learning, and even the choice of hardware determines which regularities can be exploited. Regularities are so ubiquitous in robotics that we use them without thinking about it.

Our main argument to explain the empirical results reported in this paper is the following. Active interconnections enable a more versatile composition of regularities in behavior generation when compared to passive interconnections. Since robustness stems from the ability to exhibit appropriate behavior in the presence of environmental variations, the ability to vary the behavior more richly enables robustness.

The conventional way of designing robotic systems with a high degree of modularity and simple interfaces between them leads to a simple form of composition. The interfaces are simple and passive and therefore limit how regularities can be composed. In contrast, active interconnections enable more complex compositions. They can adapt the information they pass back and forth based on the state of all of the components they connect, enabling new kinds of active compositions that are responsive to the environment as reflected in the state of the system.

We use an analogy to illustrate our argument. Let us regard regularities as basis vectors of the space of behaviors. Behavior results, as before, from the composition of these basis vectors. In this analogy, highly modular systems might represent linear combinations of the basis vectors. In contrast, active interconnections enable compositions based on any function. They can cover the space of behaviors in much more interesting ways.

Given the same regularities in a highly modular system and in a system with active interconnections, we believe the following to be true: A robotic system with active interconnections has more ways of responding to variability in the world because it can compose regularities in a more versatile manner.  Such robotic system can also be more skilled at selecting from this richer set of behaviors because this selection can be based on information about several system components. The result is that active interconnections contributes to robustness in robotic systems.

\subsection{Engineering Practice in Building Systems with Active Interconnections}

If our analysis is correct, we can increase robustness by adding more regularities to our system and by composing them more appropriately via active interconnections. This offers an intriguing perspective on designing robotic systems: designing active interconnections rather than overly sophisticated individual components. To apply this design principle, we must carefully consider the way we collaborate in system building.

From our experiences, we argue that collaborative development is crucial in designing systems with active interconnections. In particular, developers should frequently communicate about each component's limitations and challenges. This frequent knowledge exchange fosters the developer in attaining a good understanding of the bottlenecks and issues faced by the whole system. This shared understanding enables the developers to implement complex and robust robotic behaviors by establishing active interconnections between components. 

Notably, the proposed collaborative development with frequent knowledge exchange contradicts the standard software engineering practice, which suggests designing well-defined interfaces in the beginning design stage so that developers can focus on individual components, purposely minimizing communication with developers of other modules. The discrepancy arises from different design purposes: standard software engineering practice optimizes the development process against errors introduced by developers, while our engineering practice aims to enhance robustness against environmental changes by adding regularities via advantageous interconnections among system components, which requires a shared understanding about the whole system among developers. Notably, this collaborative method aligns with previous research~\cite{christensen2010cognitive}, highlighting the importance of early and continuous integration of ideas.

\subsection{Future Directions in Designing Active Interconnections}

Currently, our system's components and active interconnections are hand-engineered. Applying the proposed design principle to other tasks requires significant effort to understand the task, identify relevant patterns, and design suitable connections between components. As task complexity increases, this manual design process becomes increasingly intricate. Therefore, it is desired to use data-driven approaches to facilitate the design of robotic systems with data from either human demonstrations~\cite{ravichandar2020recent} or autonomous explorations~\cite{kroemer2021review, singh2022reinforcement}.

Interestingly, in an end-to-end learning framework, system components are jointly optimized through data, achieving comprehensive interconnectivity with minimal information loss. Such learning systems seem capable of automatically identifying and exploiting task-relevant regularities, thereby enhancing system robustness. However, these end-to-end approaches, while embodying the principle of active interconnections, often struggle to efficiently identify suitable regularities from limited data. Moreover, the regularities learned may lack the desired level of abstraction, leading to overfitting and reduced robustness to out-of-distribution scenarios. 

A promising intermediate step is to introduce the structure of active interconnections while relying on data to extract the necessary parameters for instantiation. This approach allows us to encode prior knowledge about the task structure, thereby facilitating the learning process. This is exemplified by our \emph{Guided Exploration} behavior, which employs an attention mechanism to learn regularities in interlocking dependencies from online interaction experiences. Other examples can be found in~\cite{martin2022coupled, battaje2024information} where the structure of active interconnections is based on recursive estimators, allowing the prediction and measurement models to be learned or refined from data.

\section{Conclusion}

We present a design principle for building robust robotic systems capable of handling environmental variations. This design principle is inspired by the robustness observed in biological systems and emphasizes the importance of active interconnections between system components. To evaluate the principle's effectiveness, we built robotic systems with different amounts of active interconnections. We tested their performance in manipulating lockboxes under significant environmental changes, including variations of the lockbox in scale, interlocking dependencies, joint types, relative lockbox/robot poses, and morphologies of the robot's end-effector. Our experimental results support that systems with higher amounts of active interconnections among components exhibit greater robustness under environmental variations. While this design principle has been only demonstrated in one task, we believe that our scientific arguments, examples from biological systems, empirical evidence, and system limitations showcase the potential of applying this principle to achieve robustness in a broader range of manipulation tasks. Undoubtedly, the proposed principle is not the sole pattern contributing to the robustness of biological systems. We hope this work stimulates interdisciplinary collaborations between biologists and roboticists to explore additional principles of biological robustness.

\bibliographystyle{IEEEtran} 
\bibliography{references}

\end{document}